\documentclass[runningheads]{llncs}

\usepackage{eccv}

\usepackage{eccvabbrv}

\usepackage{graphicx}
\usepackage{booktabs}
\usepackage{multirow}

\usepackage[accsupp]{axessibility}  % Improves PDF readability for those with disabilities.
\usepackage{wrapfig}

\usepackage[pagebackref,breaklinks,colorlinks,citecolor=eccvblue]{hyperref}
\usepackage[normalem]{ulem}
\usepackage{orcidlink}

\begin{document}

% \title{MineDreamer: Chain-of-Imagination Guides Sequential Decision Making in Minecraft} 
\title{MineDreamer: Learning to Follow Instructions \\ via  Chain-of-Imagination  for \\Simulated-World Control } 
% \title{MineDreamer: Chain-of-Imagination Guides Instructable Agent for Simulated-World Control} 

\titlerunning{MineDreamer}

% Second Author\inst{2,3}\orcidlink{1111-2222-3333-4444} \and
% Third Author\inst{3}\orcidlink{2222--3333-4444-5555}}

\author{
Enshen Zhou$^{1,2\ast}$, 
Yiran Qin$^{1,3\ast}$, \\
Zhenfei Yin$^{1,4}$, 
Yuzhou Huang$^{3}$, 
Ruimao Zhang$^{3\dagger}$, 
Lu Sheng$^{2\dagger}$,\\
Yu Qiao$^{1}$,
Jing Shao$^{1\ddagger}$\\
}

\authorrunning{E.~Zhou et al.}

\institute{Shanghai Artificial Intelligence Laboratory \and Beihang University \and
The Chinese University of Hong Kong, Shenzhen (CUHK-Shenzhen) \and The University of Sydney\\
\email{zhouenshen@buaa.edu.cn~~yiranqin@link.cuhk.edu.cn}\\
\large
\url{https://sites.google.com/view/minedreamer/main}
}

\maketitle

\definecolor{highlight}{HTML}{ee6002}

\newcommand{\mname}{\textsl{MineDreamer}}
\newcommand{\todo}[1]{{\color{red}#1}}
\newcommand{\update}[1]{{\color{orange}#1}}
\newcommand{\highlight}[1]{{\color{highlight}#1}}

\begin{figure*}[t]
\centering
\includegraphics[width=1\linewidth]{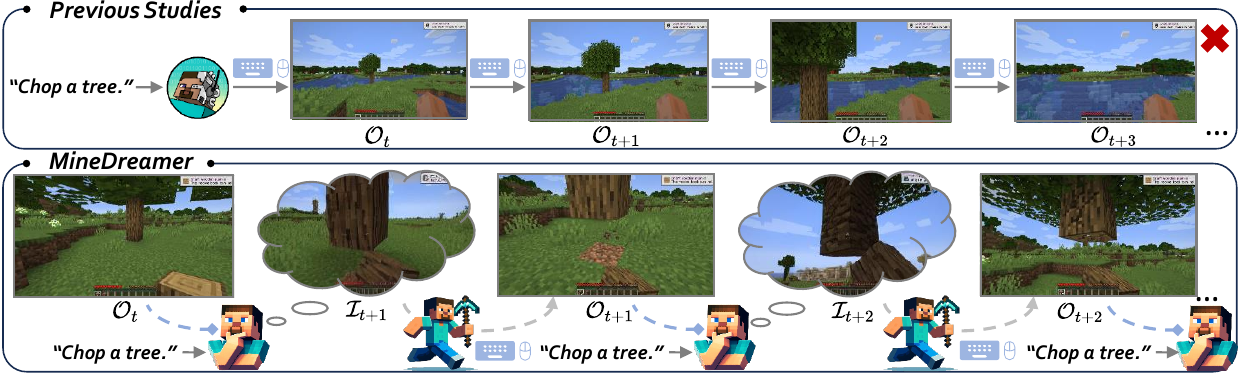}
\vspace{-6mm}
   \caption{\textbf{Comparison between {\mname} and previous studies.} In ``\texttt{Chop a tree}''~\raisebox{-0.4ex}{\includegraphics[width=0.3cm]{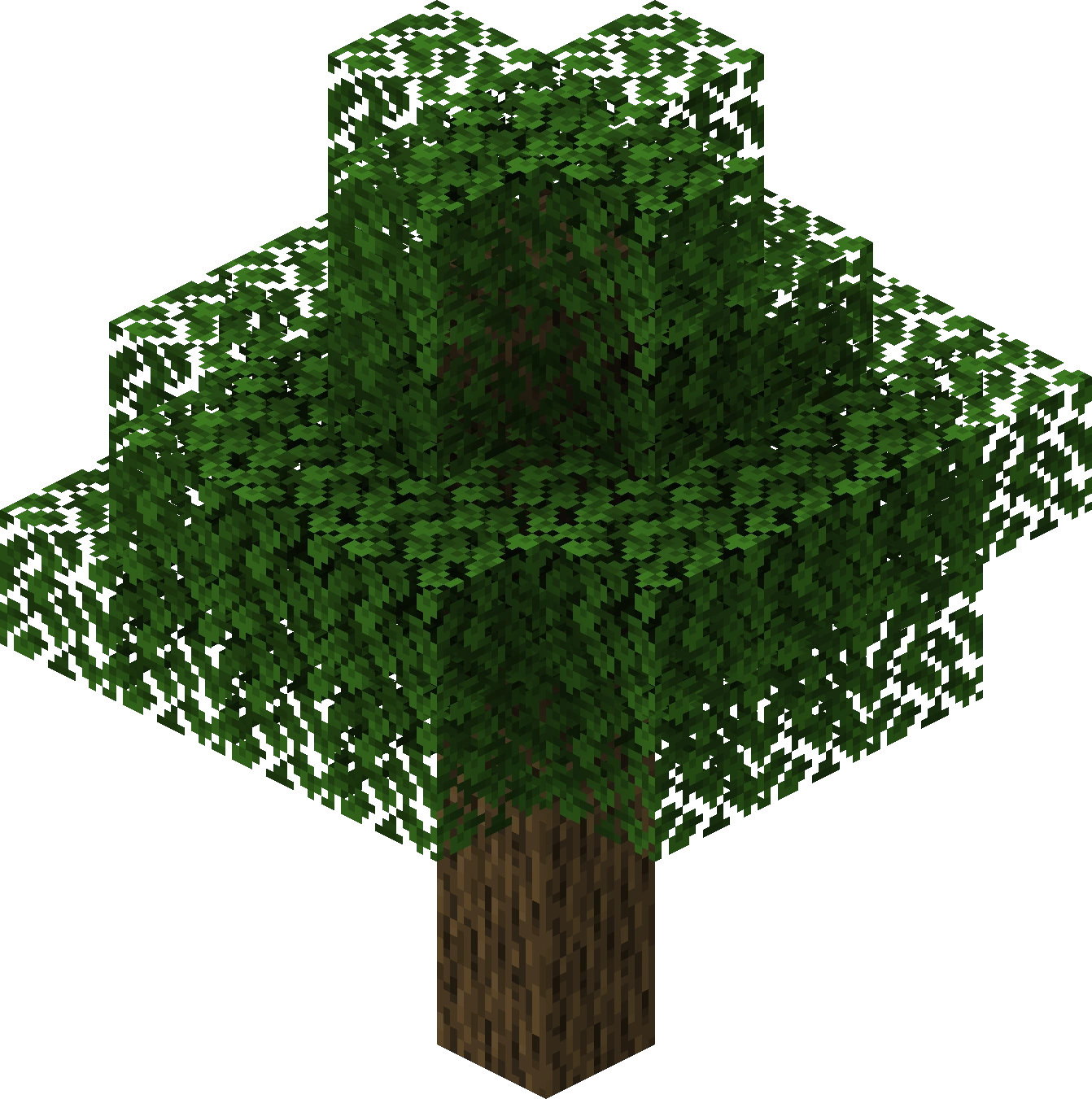}} task, {\mname} employs a Chain-of-Imagination mechanism, where it imagines step by step what to do next tailored to the current state. Imaginations contain environmental understanding and physical rules (\eg, perspective-based size changes). These can serve as more precise visual prompts to steadily guide the agent in generating actions to achieve these imaginations as effectively as possible at each step. Previous approaches have seen a tree, but missed the opportunity to chop it down.}
\vspace{-7.5mm}
\label{fig:motivation}
\end{figure*}

\let\thefootnote\relax\footnotetext{$^*$ Equal contribution\hspace{3pt} \hspace{5pt}$^\dagger$ Corresponding author\hspace{5pt} $^\ddagger$ Project leader
}
\vspace{-5mm}

\begin{abstract}
It is a long-lasting goal to design a generalist-embodied agent that can follow diverse instructions in human-like ways. 
However, existing approaches often fail to steadily follow instructions due to difficulties in understanding abstract and sequential natural language instructions. 
To this end, we introduce {\mname}, an open-ended embodied agent built upon the challenging Minecraft simulator with an innovative paradigm that enhances instruction-following ability in low-level control signal generation.
Specifically,  {\mname} is developed on top of recent advances in Multimodal Large Language Models (MLLMs) and diffusion models, and we employ a Chain-of-Imagination~(CoI) mechanism to envision the step-by-step process of executing instructions and translating imaginations into more precise visual prompts tailored to the current state;
subsequently, the agent generates keyboard-and-mouse actions to efficiently achieve these imaginations, steadily following the instructions at each step.
Extensive experiments demonstrate that {\mname} follows single and multi-step instructions steadily, significantly outperforming the best generalist agent baseline and nearly doubling its performance.
Moreover, qualitative analysis of the agent's imaginative ability reveals its generalization and comprehension of the open world.
%
% \footnote{Demo videos and imagination results are hosted on  \href{https://sites.google.com/view/minedreamer-demo}{ anonymous project webpage}.}
%

\end{abstract}
 
\vspace{-7mm}
\section{Introduction}
\label{sec:intro}

One of the core objectives of current embodied intelligence is to develop a {generalist low-level control} agent that can follow diverse instructions to solve endless open-world embodied tasks~\cite{brohan2022rt, lifshitz2023steve, cai2023groot, reed2022generalist, brohan2023rt}.
Recent studies~\cite{brohan2022rt,brohan2023rt,lifshitz2023steve, cai2023groot} successfully unlock the instruction-following ability of foundation models~\cite{baker2022video, chowdhery2023palm,chen2023pali} in the sequential decision-making domain~\cite{wen2023large,janner2021offline,chen2021decision,vinyals2019grandmaster,silver2016mastering,cai2023groot,reed2022generalist}. 
% {both in the real world and simulators}.
%
However, these methods~\cite{brohan2022rt, lifshitz2023steve} struggle to enable agents to follow textual instructions steadily, due to the:
\textbf{(1)} Many textual instructions are abstract for low-level control and models struggle to effectively understand. They should be transformed into more effective prompts that consider how to execute instructions based on the current state. Hence, simple textual instructions cannot provide a precise demonstration of the desired behavior.
\textbf{(2)} Many textual instructions are {sequential}, and executing them may require considering the current state and breaking down the task into multiple stages for step-by-step completion. Therefore, steady action generation driven by single-text instructions often fails.

To address the above issues, this work aims to explore how to unlock the situation-aware reasoning ability for a pre-trained decision-making foundation model. 
We introduce a simple yet effective mechanism called Chain-of-Imagination (CoI), which enables the agent to imagine and act upon the next stage step by step according to the instructions.
Our method is motivated by two ideas:
\textbf{(1)} When solving complex problems, humans often envision the goal of the next stage based on the current state. If we can break down the sequential instructions into multiple stages according to the current state, step by step, we can enable agents to follow instructions steadily.
%
% solving
\textbf{(2)} Inspired by prompt tuning~\cite{jia2022visual, zhou2022learning, zhou2022conditional}, if we can provide visual prompts containing physical rules and environmental understanding for each imagined step, tailored to optimally describe the desired behavior in the current state, which are more intuitive and efficient than task instructions, we can better guide the foundation model in predicting actions.
%
% Visual Prompt Tuning in computer vision offers more effective prompts tailored for foundation models in downstream tasks. If we apply the concept to train a model that generates optimal visual prompts containing physical rules and environmental understanding tailored to the current state, we can better guide the foundation model without altering its parameters.

%
To this end, we propose {\mname} within Minecraft, which generates a series of ``imagined'' sub-steps based on the textual instructions and current state. 
These visual sub-steps are then fed into a pre-trained decision-making foundation model to generate low-level control actions aimed at achieving the sub-steps.
Specifically, {\mname} comprises three modules:
\textbf{(1)} {An Imaginator, a diffusion model enhanced by a Multimodal Large Language Model (MLLM), can better generate imaginations that contain the physical rules and environmental understanding.}
\textbf{(2)} A Prompt Generator, the bridge between Imaginator and PolicyNet, can convert future imaginations into latent visual prompts that offer more logical and precise demonstrations of the desired behavior.
\textbf{(3)} A PolicyNet, a foundation model, can use latent prompts as guidance to predict actions for agents in an open-world environment.

Notably, as shown in \cref{fig:motivation}, {\mname} leverages a Chain-of-Imagination mechanism through multi-turn interaction between the Imaginator and the PolicyNet and cyclically generates latent visual prompts that better align with the current state to guide the PolicyNet in following instructions steadily in action generation.
%
% designed to sequentially imagine the steps to complete tasks based on current observations. 
%
%
This mechanism represents an attempt to implement ``self multi-turn interaction'' in the sequential decision-making domain.
Training an Imaginator in an open-world environment to envision the image of the next step requires extensive data. 
We employ the \textit{Goal Drift Collection} method to gather a large amount of egocentric embodied data, which helps the Imaginator to understand how to achieve the instruction sequentially and how to achieve it repeatedly.

% %
% %
% Extensive experiment prove the {\mname} can reliably follow human instructions to accomplish tasks required for single- or multi-step instructions, outperforming the best baseline by nearly two times on the evaluation metrics.
% %
% Our qualitative analysis of goal imaginations generated by Dreamer further reveals some interesting phenomena of the generative model, which allows for a certain degree of understanding of the physical rules in an open world.

Our main contributions are as follows: 
\begin{itemize}
    \item We introduce the Chain-of-Imagination(CoI) method, {which introduces ``self multi-turn interaction'' to the sequential decision-making domain} and enables the agent to follow human instructions steadily in action generation.
    \item We propose the \textit{Goal Drift Collection} method and an MLLM-enhanced diffusion model that can generate imaginations adhering to physical rules and environmental understanding, providing more precise visual prompts relevant to the current state and instructions.
    % We propose the \textit{Goal Drift Collection} method and MLLM-enhanced diffusion model that can generate imaginations that reflect physical rules and environmental knowledge, serving as more precise visual prompts tailored to the current state.
    \item Leveraging these methods, we create an embodied agent in Minecraft named {\mname} that has achieved nearly double the performance of the best generalist agent baseline in executing single and multi-step instructions steadily.
    % Leveraging the methods above, we developed a Minecraft embodied agent called {\mname} that has achieved nearly double the performance of the best generalist agent baseline in executing single and multi-step instructions.
    
    % \item We will release model weights, training scripts, and evaluation code to promote further research in creating generative models for diffusion-based open-ended sequential low-level control agents.
\end{itemize}
\vspace{-3mm}
\section{Related Work}
\label{sec:related work}

\subsection{Build Instruction-Following Agents in Minecraft}
Research on generalist agents in Minecraft's complex and dynamic environment is increasingly popular in AI. 
Despite the exploration of Large Language Models~\cite{radford2019language,brown2020language,ouyang2022training,touvron2023llama,touvron2023llama2,chiang2023vicuna} as high-level task planners that guide agents in executing long-horizon tasks~\cite{qin2023mp5,wang2023jarvis,wang2023describe,wang2023voyager,zhu2023ghost, gong2023mindagent} like Voyager~\cite{wang2023voyager} and MP5~\cite{qin2023mp5}, we still require lower-level controllers~\cite{hafner2023mastering, baker2022video, cai2023groot, lifshitz2023steve, fan2022minedojo} to execute the generated plans.
In the sequential decision-making domain, DreamerV3~\cite{hafner2023mastering} trains agents using a world model, while VPT~\cite{baker2022video} builds a large foundational model to generate actions by learning from extensive video data. However, neither can follow instructions. 
GROOT~\cite{cai2023groot} is developed to follow video instructions but fails to follow text instructions.
STEVE-1~\cite{lifshitz2023steve}, an evolution of VPT~\cite{baker2022video}, is built for text instructions but struggles to understand natural language prompts, despite extensive prompt engineering.
Therefore, we create {\mname}, which, leveraging the Chain-of-Imagination mechanism, generates more precise visual prompts step-by-step, enabling it to follow instructions steadily in action generation.

\vspace{-4mm}
\subsection{Conditioned Diffusion Models in Embodied Scenario}
With the development of the text-to-image diffusion model~\cite{dhariwal2021diffusion, ho2022classifier, nichol2021glide, ramesh2022hierarchical, saharia2022photorealistic, rombach2022high}, the instruction-based diffusion methods~\cite{zhang2023magicbrush, brooks2023instructpix2pix, cao2023masactrl, kawar2023imagic, tumanyan2023plug, hertz2022prompt, geng2023instructdiffusion,fu2023guiding,huang2023smartedit} have recently marked considerable progress in generative tasks, especially in embodied scenarios.
UniPi~\cite{du2024learning} and HiP~\cite{ajay2024compositional} integrate video diffusion with inverse dynamics to generate robot control signals for specific tasks.
SkillDiffuser~\cite{liang2023skilldiffuser} applies interpretable hierarchical planning via skill abstractions in diffusion-based task execution.
While existing methods can only handle embodied tasks limited to fixed environments, 
the emergence of 
% Large Language Models(LLMs)~\cite{radford2019language,brown2020language,ouyang2022training,touvron2023llama,touvron2023llama2,chiang2023vicuna} and 
Multimodal Large Language Models~(MLLMs)~\cite{liu2023visual,yin2023lamm,zhu2023minigpt,ye2023mplug,gao2023llama,peng2023kosmos,shi2023chef,chen2023octavius} has showcased superior reasoning and perceptual abilities in open-world environment. 
Inspired by this, we create an MLLM-enhanced diffusion model, focusing on the model's understanding of physics rules and environmental understanding, and its ability to create high-quality egocentric images for guiding low-level action generation.

\vspace{-4mm}
\section{Method}
\vspace{-2mm}

In this section, we first provide an overview~(\cref{sub:overview}) of our {\mname}, including its mechanisms and features.
Next, we introduce the purpose and workflow of the Chain-of-Imagination~(CoI) mechanism~(\cref{sub:coi}) regarding \cref{fig:pipeline}.
To implement CoI and collect extensive embodied data to train Imaginator, we elaborate on the dataset construction~(\cref{sub:datasets}), including \textit{Goal Drift Collection} method.
Finally, we provide the necessary details of each part, including Imaginator (\cref{sub:imaginator}), Prompt Generator, and PolicyNet~(\cref{sub:policy}).

\label{sec:Method}

\vspace{-4mm}
\subsection{Overview}
\label{sub:overview}
Our {\mname} comprises three modules, \ie, Imaginator, Prompt Generator, and PolicyNet. 
Our objective is to empower agents, especially foundation models in the sequential decision-making domain, to follow human instructions steadily and act accordingly. 
The Imaginator is a parameter-efficiently fine-tuned diffusion model specific to Minecraft utilizing the visual reasoning ability of a Multimodal Large Language Model (MLLM). 
The Prompt Generator reconstructs latent visual prompts from the current observations, future imaginations, and instructions. 
PolicyNet is the existing Video Pretraining (VPT)~\cite{baker2022video} model, trained on 70k hours of Minecraft gameplay.

\vspace{-4.5mm}
\subsubsection{Why future goal imagination?}
\label{subsub:Why future goal imagination}
% \vspace{+1mm}
% \noindent{\textbf{Why future goal imagination? }}
Given a pre-trained model that can predict actions, the intuitive approach is to input the current state and instructions to guide it directly. So why the future goal imagination? 
In practice, we find that future goal imagination proves more interpretable for humans, easing debugging, and improving interaction and safety assessment~\cite{zhang2024psysafe, li2024salad, qu2023unsafe,rando2022red}.
Furthermore, images yield flexible, explicit representations, facilitating natural language goal decomposition into clearer stages by learned physical rules and environmental understanding, helping the low-level control model ``plan'' what to do now.

\vspace{-4.5mm}
\subsubsection{Why can {\mname} follow instructions more steadily?}
\label{subsub:Why can model follow instructions more steadily}
% \vspace{+1mm}
% \noindent{\textbf{Why can {\mname} follow instructions more steadily? }}
Firstly, {\mname} employs a Chain-of-Imagination (CoI) mechanism for incremental goal achievement via self-multi-turn interactions, enabling the agent to appropriately respond to the current state.
In addition, with the help of this mechanism, the Prompt Generator crafts logical latent visual prompts that provide clear demonstrations of desired behaviors, ensuring that the agent steadily follows instructions.
Furthermore, the enhanced Imaginator not only comprehends open-ended visual concepts, enabling it to imagine images of novel instructions it has never seen before but also ensures these images adhere to physical rules and environmental understanding, thereby sharpening the precision of prompts.
Thus, {\mname} can follow instructions steadily in an open-world environment.

\begin{figure*}[t]
\centering
\includegraphics[width=1\linewidth]{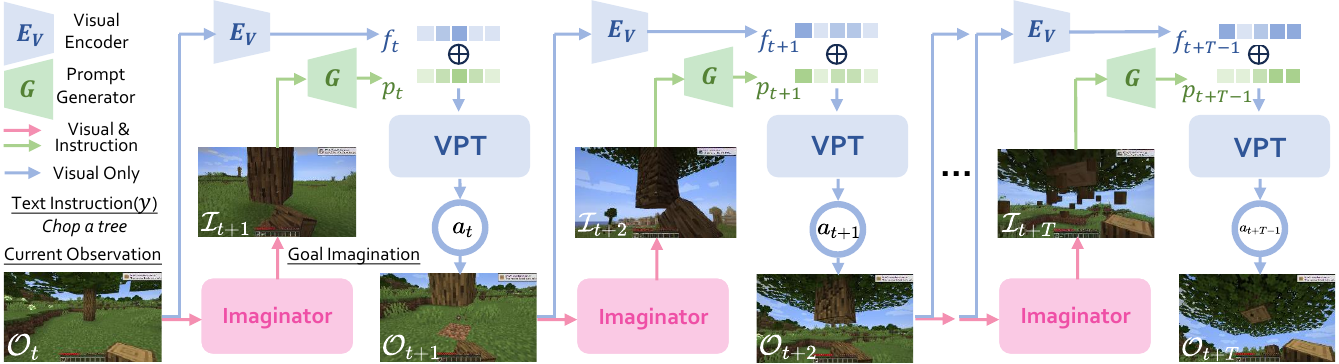}
\vspace{-5.5mm}
   \caption{\textbf{The Overview of Chain-of-Imagination.} The Imaginator imagines a goal imagination based on the instruction and current observation. The Prompt Generator transforms this into a precise visual prompt, considering both the instruction and observed image. The Visual Encoder encodes the current observation, integrates it with this prompt, and inputs this into VPT. VPT then determines the agent's next action, leading to a new observation, and the cycle continues. Note that VPT's input is historical observations, so the figure cannot fully represent the autoregressive process. More details about VPT as PolicyNet can be found in Sec.~\ref{sub:policy}.}
\vspace{-7.5mm}
\label{fig:pipeline}
\end{figure*}

\vspace{-4mm}
\subsection{Chain-of-Imagination}
\label{sub:coi}
\vspace{-1mm}
Chain-of-Imagination (CoI) enables the agent to envision the steps needed to achieve a goal iteratively. As shown in \cref{fig:pipeline}, it is an example to demonstrate how CoI works.
First, the Imaginator takes in the user's instructions $y$ and current observations $\mathcal{O}_{t}$ and imagines a future image $\mathcal{I}_{t+1}$ depicting a moment within the process of completing the given instruction $y$, which is closely related to the current observation $\mathcal{O}_{t}$.
Next, the Prompt Generator progressively creates a more precise latent visual prompt $p_{t}$ in awareness of the current observation $\mathcal{O}_{t}$, instruction $y$ and future imagination $\mathcal{I}_{t+1}$, aligning with the visual input space of the Video Pretraining (VPT)~\cite{baker2022video} model.
The Visual Encoder then processes $\mathcal{O}_{t}$ into a representation $f_{t}$, which is combined with $p_{t}$ and fed into VPT~\cite{baker2022video}.
Finally, VPT~\cite{baker2022video} progressively predicts an action~(\ie, keyboard and mouse) from the observation history, interacts with the environment, gathers a new observation $\mathcal{O}_{t+1}$, and repeats the cycle later.

\vspace{-4mm}
\subsection{Datasets}
\label{sub:datasets}
\vspace{-1mm}
We train the Imaginator with the Goal Drift Dataset, which includes 500k triplets (current observation, future goal imagination, instruction) from the OpenAI Contractor Gameplay Dataset~\cite{baker2022video}, using the \textit{Goal Drift Collection} method.

\vspace{-5.5mm}
\subsubsection{OpenAI Contractor Gameplay Dataset.}
\label{subsub:OpenAI Contractor Gameplay Dataset}
% \vspace{+1mm}
% \noindent{\textbf{OpenAI Contractor Gameplay Dataset. }}
OpenAI Contractor Gameplay Dataset~\cite{baker2022video} is created by hiring human contractors to play Minecraft and complete tasks like house building.
Game events, like ``\texttt{mine\_block}'', noting the type of block broken, are logged with timestamps. These timestamps ($t^*$) provide precise progress tracking and align with completed event-related instructions.
\vspace{-5.5mm}
\subsubsection{Goal Drift Collection.}
\label{subsub:Goal Drift Dataset Collection}
% \vspace{+1mm}
% \noindent{\textbf{Goal Drift Dataset Collection. }}
The Gameplay Dataset allows us to construct numerous embodied data by using specific event-related instructions achieved at each timestamp $t^*$. 
% \footnote{\url{https://github.com/openai/Video-Pre-Training}}
Yet, directly pairing images from these timestamps $t^*$ as future goal imaginations $\mathcal{O}_{t^*}$ with images from a fixed timestep $T$ earlier as current observations $\mathcal{O}_{t^* - T}$, along with instruction $y$, could lead to certain problems:
\textbf{(1) Goal Illusion: The Imaginator edits the observation to depict the completed instruction.}
Training the Imaginator on such data may reduce it to an image editor, as it generates imaginations without regard to the environment because all goal imaginations in the dataset represent the moment when instruction is completed. For instance, given the instruction ``\texttt{Break dirt}''~\raisebox{-0.3ex}{\includegraphics[width=0.3cm]{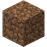}} while facing the sky, the Imaginator may unrealistically insert a broken dirt block~\raisebox{-0.3ex}{\includegraphics[width=0.3cm]{icon/dirt.png}} into the sky.
\textbf{(2) Imagination Stagnation: The Imaginator fails to conceive repeated task completion.} 
The Imaginator is trained to envision the instructions' fulfillment once, not recognizing the need for repetition, as all current observations precede the achievement of instructions.
For instance, given ``\texttt{Chop a tree}''~\raisebox{-0.4ex}{\includegraphics[width=0.3cm]{icon/tree.png}}, after cutting the uppermost wood~\raisebox{-0.3ex}{\includegraphics[width=0.3cm]{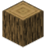}} by looking up, the agent will not look down for more trees~\raisebox{-0.4ex}{\includegraphics[width=0.3cm]{icon/tree.png}}, impeding continuous task performance.

\begin{figure*}[t]
\centering
\includegraphics[width=1\linewidth]{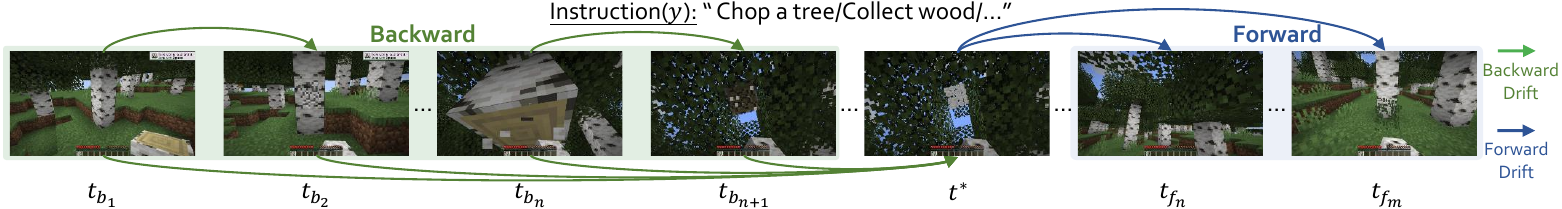}
\vspace{-5.5mm}
   \caption{\textbf{Goal Drift Collection. } For each timestamp $t^*$, we form many triplets comprising (current observation, goal imagination, instruction) associated with the game event-related instructions completed by contractors. Each pair of linked images forms a training triplet with its instruction for the Imaginator in this figure.}
\vspace{-7.5mm}
\label{fig:gd}
\end{figure*}

To address the aforementioned issues, we propose the \textit{Goal Drift Collection} method to gather Goal Drift Dataset. 
From the Gameplay Dataset, we form many triplets (current observation, goal imagination, instruction) at each timestamp $t^*$, all associated with the same event-related instructions $y$ completed by the contractors. \cref{fig:gd} shows that a pair of linked images with instructions $y$ constitutes a training triplet. Our approach has both Backward Drift, which helps the model understand the step-by-step completion of tasks to mitigate Goal Illusion, and Forward Drift, which enables the model to learn how to accomplish instructions repeatedly to reduce Imagination Stagnation.
The details of collecting three kinds of data samples corresponding to each $t^*$ are as follows:
\begin{enumerate}
\vspace{-2mm}
    \item Backward Drift 1: We set $t_{b_{1}}$ as $t^*$ backward by fixed $T_b$ time steps and then select $m-2$ random timestamps between $t_{b_{1}}$ and $t^*$ to form the sequence $t_{b_{1}},\ldots,t_{b_{m}}$ , where $t^*$ is $t_{b_{m}}$. At each time step, the current and next observations are paired as the current observations and goal imagination, respectively, which can form $m-1$ samples.
    \item Backward Drift 2: In $t_{b_{1}},\ldots,t_{b_{m}}$, the observations at each timestamp except for $t_{b_{m}}$ are used as the current observations, and the observation at $t^*$ serve as the goal imagination, which can form $m-1$ samples.
    \item Forward Drift: We set $t_{f_{m}}$ as $t^*$ forward by fixed $T_f$ time steps and randomly select $m-2$ timestamps between $t^*$ and $t_{f_{m}}$ , where $t^*$ is $t_{f_{1}}$. The observation at $t^*$ serves as the current observation, and the observations at future timestamps serve as the goal imaginations, which can form $m-1$ samples.
\vspace{-1mm}
\end{enumerate}

\noindent For more details about the dataset and collection method, please check Supp.~\ref{sup:Dataset}.

\vspace{-3mm}
\subsection{Imaginator}
\label{sub:imaginator}

\begin{figure*}[t]
\centering
\includegraphics[width=1\linewidth]{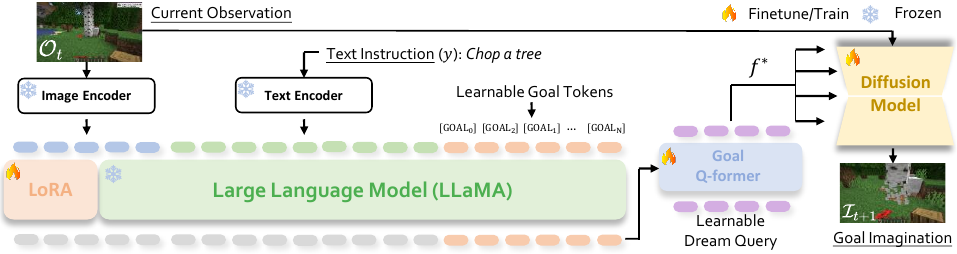}
\vspace{-6.5mm}
   \caption{\textbf{The Overall Framework of Imaginator.} For the goal understanding, we add $k$ $[\mathrm{GOAL}]$ tokens to the end of instruction $y$ and input them with current observation $\mathcal{O}_{t}$ into LLaVA~\cite{liu2023visual}. Then LLaVA~\cite{liu2023visual} generates hidden states for the $[\mathrm{GOAL}]$ tokens, which the Q-Former processes to produce the feature $f^*$. Subsequently, the image encoder $\mathbf{E}_{v}$ combines its output with $f^*$ in the diffusion models for instruction-based future goal imagination generation.}
\vspace{-7mm}
\label{fig:imaginator}
\end{figure*}

Inspired by prompt tuning~\cite{jia2022visual, zhou2022learning, zhou2022conditional}, we introduce Imaginator, an MLLM-enhanced diffusion model that imagines step by step what to do next based on the current state and instruction, enabling the creation of more precise visual prompts for improved low-level control demonstrations of the desired behavior.
Imaginator's training data utilizes the Goal Drift Dataset from Sec~\ref{sub:datasets}, consisting of (current observation, goal imagination, instruction) triplets.

\vspace{-4.5mm}
\subsubsection{Goal Understanding via Task Instruction Following.}
\label{subsub:Goal Understanding via Task Instruction Following}
Given a current observation $\mathcal{O}_{t}$ and a textual instruction $y$, the Imaginator generates a future goal imagination $\mathcal{I}_{t+1}$ for the PromptGenerator's visual prompt. In Fig.~\ref{fig:imaginator}, current observation $\mathcal{O}_{t}$ is encoded by a frozen image encoder $\mathbf{E}_{v}$ into $\mathbf{E}_{v}(O_{t})$, textual instruction $y$ is tokenized into $(x_{1}, ..., x_{T})$, they are sent to the LLM together. Imaginator now can acquire a goal imagination of the instruction intention but are limited to the language modality. 
Inspired by GILL~\cite{koh2024generating}, we bridge the language-vision modalities gap by extending the LLM's vocabulary with k Learnable Goal Tokens $[\operatorname{GOAL}_{1}], \ldots,[\operatorname{GOAL}_{k}]$, appending them to instruction $y$. 
Specifically, a trainable matrix $\mathbf{E}_{g}$, representing these $[\operatorname{GOAL}]$ embeddings, is added to the LLM's embedding matrix. 
We aim to minimize the negative log-likelihood of predicting the next $[\operatorname{GOAL}]$ token given previously generated $[\operatorname{GOAL}]$ tokens:

{
\scriptsize
\begin{equation}
\mathcal{L}_{\mathrm{LLM}}=-\sum_{i=1}^k \log p_{\left\{\theta_{L} \cup \theta_{l} \cup \mathbf{E}_{g}\right\}} ([\operatorname{GOAL}_{i}] \mid \mathbf{E}_{v}(O_{t}),\nonumber x_{1}, ..., x_{T}, [\operatorname{GOAL}_{1}], \ldots,[\operatorname{GOAL}_{i-1}])
\tag{1}
\end{equation}
}

\noindent We add LoRA~\cite{hu2021lora} parameters $\theta_{l}$ into the LLM's self-attention projection layers for efficient fine-tuning while keeping all LLM parameters $\theta_{L}$ frozen. During training, only the LoRA~\cite{hu2021lora} parameters $\theta_{l}$ and the Learnable Goal Tokens $\mathbf{E}_{g}$ are updated. The hidden states $h_{[\operatorname{GOAL}]}$ corresponding to $\mathbf{E}_{g}$ tokens are used to generate imaginations in the following module.

\vspace{-4.5mm}
\subsubsection{Goal Imagination Generation via Latent Imagination.}
\label{subsub:Goal Imagination Generation via Latent Imagination}
% \vspace{+1mm}
% \noindent {\textbf{Goal Imagination Generation via Latent Imagination. }}
To address the disparity between the LLM's hidden states and the CLIP~\cite{radford2021learning} text encoder's feature spaces, we must transform the LLM's sequential goal tokens into semantically relevant representations for guiding goal imagination generation. Inspired by BLIP2~\cite{li2023blip} and InstructBLIP~\cite{instructblip}, we employ a Goal Q-Former $\mathcal{Q}$ with several Learnable Dream Query, to derive the goal imagination representation $f^*$:

{
\vspace{-2mm}
% \small
\begin{equation}
f^*=\mathcal{Q}\left(h_{[\operatorname{GOAL}]}\right)
\tag{2}
\end{equation}
}

\noindent To enhance goal imagination with representation $f^*$ to guide imagination generation, we utilize a latent diffusion model combining a variational autoencoder (VAE)~\cite{kingma2013auto} for latent space denoising diffusion. 
Drawing from InstructPix2Pix's~\cite{brooks2023instructpix2pix} latent diffusion approach, a cornerstone in instruction-based image editing, our model introduces noise to the latent encoding $z=\mathcal{E}(\mathcal{I}_{t+1})$ of the goal imagination $\mathcal{I}_{t+1}$ through encoder $\mathcal{E}$, yielding a noisy latent $z_{s}$ across timesteps $s \in S$. A U-Net~\cite{ronneberger2015u} $\epsilon_\delta$ is trained to estimate this noise, conditional on the current observation $c_{o}=\mathcal{E}(O_{t})$ and text instruction $c_{T}$, by merging $c_{o}$ with $z_{s}$. The specific process can be formulated as follows:
{
\small
\begin{equation}
\mathcal{L}_{\mathrm{dream}}=\mathbb{E}_{\mathcal{E}(\mathcal{I}_{t+1}), \mathcal{E}(O_{t}), c_{T}, \epsilon \sim \mathcal{N}(0,1), s}[\| \epsilon \nonumber -\epsilon_\delta(s, \mathrm{concat}[z_s, \mathcal{E}(O_{t})]+f^*) \|_2^2]
\tag{3}
\label{eq:diffusion}
\end{equation}
}

\noindent where $\epsilon$ is unscaled noise, $s$ is the sampling step, $z_{s}$ is latent noise at step $s$, $\mathcal{E}(O_{t_{n}})$ is the current observation condition, and $c_{T}$ is the text instruction condition. The $\mathrm{concat}$ corresponds to the concatenation operation.

\vspace{-4mm}
\subsection{Prompt Generator and PolicyNet}
\label{sub:policy}
To transform goal imaginations into precise latent visual prompts that the PolicyNet can understand, we require a Prompt Generator to serve as the bridge between the Imaginator and the PolicyNet. 
Inspired by STEVE-1~\cite{lifshitz2023steve}, our prompt generator is a conditional variational autoencoder (CVAE)~\cite{sohn2015learning, kingma2013auto} model trained on the Goal Drift subset dataset. It encodes the current observations, goal imaginations, and instructions by MineCLIP~\cite{fan2022minedojo} to produce three embeddings. These embeddings are then reconstructed into a latent visual embedding within the MineCLIP~\cite{fan2022minedojo} visual space and a linear layer then projects it into the visual input space of our PolicyNet.

In our PolicyNet, we utilize the architecture of the existing model named VPT~\cite{baker2022video} and the training parameters of STEVE-1~\cite{lifshitz2023steve}. 
% This also includes the parameters of the linear layer from the Prompt Generator in STEVE-1~\cite{lifshitz2023steve}.
%
Specifically, as shown in \cref{fig:pipeline}, we first process the current observation with a Visual Encoder (\ie, ResNet~\cite{he2016deep}) of VPT~\cite{baker2022video} and get representation $f_{t}$. After adding it with the latent visual prompts $p_{t}$ generated by the Prompt Generator, the sum result $o_{t}$ is then fed into the PolicyNet.
PolicyNet, whose backbone is Transformer-XL~\cite{dai2019transformer}, processes the current input representations $o_{t}$ and autoregressively predicts the next action $a_{t}$.
We can describe the process where the Prompt Generator creates latent visual prompts $p_{t}$ and PolicyNet predicts the next action $a_{t}$ based on them and historical observations using the following simple notation:

\vspace{-2mm} 
{
\small
\begin{gather}
p_{t} \leftarrow \mathcal{G}(\mathcal{O}_{t}, \mathcal{I}_{t+1}, y) , ~~~ f_{t} \leftarrow \mathcal{V}(\mathcal{O}_{t}),~~~ o_{t} \leftarrow f_{t} + p_{t}, ~~~ {a}_{t} \leftarrow \mathcal{T}(o_{t-T}, \ldots, o_{t})\tag{4}
\end{gather}
}

\noindent where $\mathcal{G}$ is PromptGenerator, $\mathcal{V}$ is VisualEncoder, and $\mathcal{T}$ is TransformerXL~\cite{dai2019transformer}.
\vspace{-3mm}
\section{Experiments}
\vspace{-2mm}
\subsection{Experimental Setup}
\subsubsection{Training Process.}
\label{subsub:Training Process}
% \vspace{+1mm}
% \noindent{\textbf{Training Process. }}
The training process of Imaginator is divided into three main stages. In the first stage, the MLLM is aligned with the CLIP~\cite{radford2019language} text encoder~\cite{radford2021learning} using the QFormer~\cite{li2023blip}. In the second stage, we apply InstructPix2Pix~\cite{brooks2023instructpix2pix} to warm up the weights for the diffusion model in Minecraft. In the third stage, we optimize Imaginator in an end-to-end manner. To be specific, the weights of LLaVA~\cite{liu2023visual} are frozen and LoRA~\cite{hu2021lora} is added for efficient fine-tuning. For the diffusion model, we directly use the weights pre-trained in the second stage as the initial weights in Imaginator. 
The CVAE~\cite{sohn2015learning, kingma2013auto} within the Prompt Generator features a Gaussian prior and a Gaussian posterior, with its encoder and decoder, parameterized as three-layer MLPs, each with 512 hidden units and layer normalization~\cite{ba2016layer}, similar to the architecture of STEVE-1's~\cite{lifshitz2023steve} prior.
More training details can be found in Supp.~\ref{sup:Implementation}.
%
% \vspace{-5.5mm}
% \subsubsection{Implementation Details.}
% \label{subsub:Implementation Details}
% % \noindent \textbf{Implementation Details.}
% For the Large Language Model with visual input (e.g., LLaVA), we choose LLaVA-1.1-7b as the base model. During training, the weights of LLaVA are frozen and we add LoRA for efficient fine-tuning. We expand the original LLM vocabulary with $32$ new tokens. The QFormer is composed of $6$ transformer~\cite{vaswani2017attention} layers and $77$ learnable query tokens. 

% We use the AdamW optimizer~\cite{loshchilov2017decoupled} in all three stages. 
% In the initial stage of training, we configure the learning rate and weight decay parameters at 2e-4 and 0, respectively. The training targets for this stage encompass a dual-objective framework, comprising the Mean Squared Error~(MSE) loss between the outputs of LLaVA and the clip text encoder, alongside the language model loss. Both losses are assigned equal weights of 1.
% The training setting in the second is the same as InstructPix2Pix~\cite{brooks2023instructpix2pix}.
% In the final stage, the settings for the learning rate, weight decay, and warm-up ratio are adjusted to 1e-5, 0, and 0.001, respectively. During this phase, the loss function is the diffusion loss. 
%
\vspace{-5.5mm}
\subsubsection{Training Datasets.}
\label{subsub:Training Datasets}
% \vspace{+1mm}
% \noindent \textbf{Training Datasets.
In the first stage of Imaginator, we use the extensive corpus CC12M~\cite{changpinyo2021conceptual}, and our Goal Drift Dataset is used in the second and third stages. 
We follow STEVE-1's~\cite{lifshitz2023steve} approach for CVAE~\cite{sohn2015learning, kingma2013auto} training, curating a subset of approximately 10k quadruplets from the Goal Drift Dataset for our test tasks. This subset includes current observations, goal imaginations, and instructions that match the Goal Drift Dataset. We use the MineCLIP~\cite{fan2022minedojo} video encoder to transform the goal imagination and the previous 16 frames into a visual prompt embedding, which acts as the ground truth. More details can be found in Supp.~\ref{sup:Dataset}.

\vspace{-5.5mm}
\subsubsection{Environment Setting.}
\label{subsub:Environment Setting}
% \vspace{+1mm}
% \noindent{\textbf{Environment Setting.}}
We employ MineRL~\cite{guss2019minerl} as the Minecraft simulation. The observation space is limited to RGB images, and the action space is confined to keyboard and mouse controls, which are consistent with human interaction. For more details about the simulator, please check Supp.~\ref{sup:Environment}.

\vspace{-5.5mm}
\subsubsection{Baseline.}
\label{subsub:Baseline}
% \vspace{+1mm}
% \noindent{\textbf{Baseline.}}
We compare {\mname} with three baseline: 
\begin{enumerate}
    \item VPT~\cite{baker2022video}, a foundation model pretrained on 70k hours gameplay. Here, we select the VPT(rl), which is finetuned by reinforcement learning on the original VPT~\cite{baker2022video} foundation model but \textbf{cannot follow instructions}.
    \item STEVE-1~\cite{lifshitz2023steve}, an instruction-following agent finetuned from VPT(rl). Here, we select STEVE-1(text), which uses a simple prior to aligning the text with the visual space,\textbf{ without considering the current observation}.
    \item Multi-Modal Memory, a substitute for the Imaginator and Prompt Generator in {\mname}, efficiently searches through extensive instruction-video pairs to find the most relevant video as a visual prompt based on the given instruction and the current observation, which effectively \textbf{leverages the current observation and incorporates a CoI mechanism}.    
\end{enumerate}
For more details about the baseline, please check Supp.~\ref{supsub:Baseline Datails}.
 
\vspace{-5.5mm}
\subsubsection{Evaluation.}
\label{subsub:Evaluation}
% \vspace{+1mm}
% \noindent{\textbf{Evaluation.}}
We utilize STEVE-1's~\cite{lifshitz2023steve} \textit{early-game evaluation suite}, which comprises two evaluations: 
\textbf{(1)} Programmatic Evaluation, a quantitative evaluation used to evaluate an agent's ability to execute \textbf{single-step} instruction steadily. We track the states provided by the simulator to calculate metrics (\eg, wooden log collection, travel distance).
\textbf{(2)} Command-Switching Evaluation, a quantitative evaluation designed to assess whether the agent can successfully execute \textbf{multi-step} instructions in sequence to complete long-horizon tasks (\eg, obtaining diamond~\raisebox{-0.3ex}{\includegraphics[width=0.3cm]{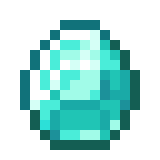}}). We use the success rate as the metric for evaluation.
More evaluation details can be found in Supp.~\ref{supsub:Programmatic} and Supp.~\ref{supsub:Command-Switching}.

\vspace{-4mm}
\subsection{Performance on Textul Instructions Control}

\begin{figure*}[t]
\centering
\includegraphics[width=1\linewidth]{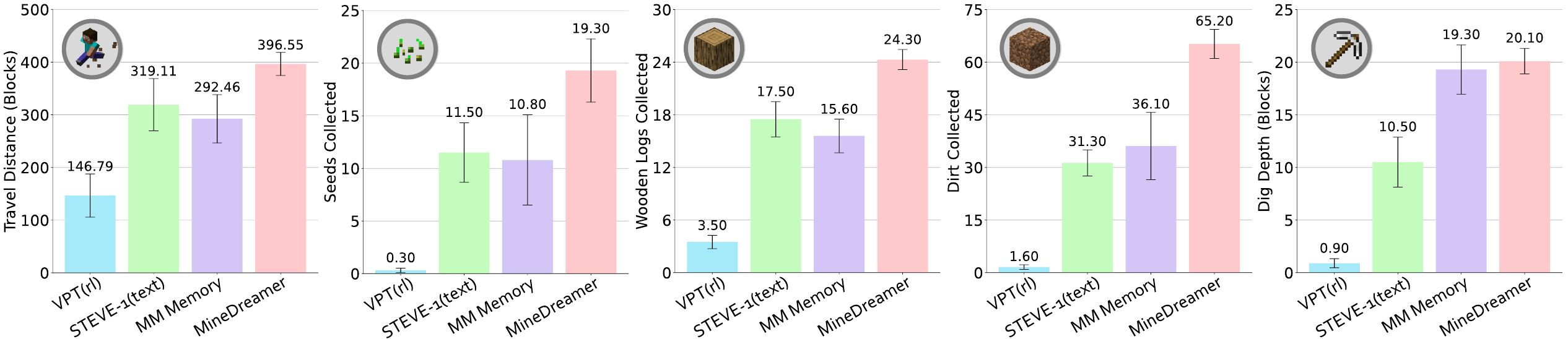}
\vspace{-5.5mm}
   \caption{\textbf{Performance on Programmatic Evaluation.} {\mname} surpasses the unconditional VPT~\cite{baker2022video}, the text-conditioned STEVE-1~\cite{lifshitz2023steve} that ignores current state, and the Multi-Modal Memory that utilizes current state with a CoI mechanism.}
\vspace{-5.5mm}
\label{fig:programmatic_evaluation}
\end{figure*}

% \vspace{-5mm}
% \noindent{\textbf{Programmatic Evaluation. }} 
\subsubsection{Programmatic Evaluation.}
\label{subsub:Programmatic Evaluation}
We quantitatively evaluate all agents on 5 tasks and plot the programmatic metric performances(mean and 95\% confidence intervals). 
Each task runs 10 trials with distinct environment seeds, limiting 3,000 frames (\ie, 2.5 minutes of gameplay) which are consistent with STEVE-1~\cite{lifshitz2023steve}. Unlike STEVE-1~\cite{lifshitz2023steve}, we condition all agents with the most suitable biome.

\cref{fig:programmatic_evaluation} compares the performance of our {\mname} with the unconditional VPT~\cite{baker2022video}, the text-conditioned STEVE-1~\cite{lifshitz2023steve} and {\mname} using Multi-Modal Memory. With appropriate text instructions, {\mname} significantly outperforms the unconditional VPT~\cite{baker2022video}, collecting 64$\times$ more seeds~\raisebox{-0.3ex}{\includegraphics[width=0.3cm]{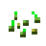}}, 7$\times$ more wood~\raisebox{-0.3ex}{\includegraphics[width=0.3cm]{icon/wood.png}}, 41$\times$ more dirt~\raisebox{-0.3ex}{\includegraphics[width=0.3cm]{icon/dirt.png}}, traveling 2.7$\times$ further~\raisebox{-0.3ex}{\includegraphics[width=0.3cm]{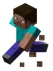}}, and digging 22$\times$ deeper~\raisebox{-0.3ex}{\includegraphics[width=0.3cm]{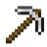}}. It also surpasses the STEVE-1~\cite{lifshitz2023steve}, collecting 1.7$\times$ more seeds~\raisebox{-0.3ex}{\includegraphics[width=0.3cm]{icon/seed.png}}, 1.4$\times$ more wood~\raisebox{-0.3ex}{\includegraphics[width=0.3cm]{icon/wood.png}}, 2.1$\times$ more dirt~\raisebox{-0.3ex}{\includegraphics[width=0.3cm]{icon/dirt.png}}, traveling 1.2$\times$ further~\raisebox{-0.3ex}{\includegraphics[width=0.3cm]{icon/explore.png}}, and digging 1.9$\times$ deeper~\raisebox{-0.3ex}{\includegraphics[width=0.3cm]{icon/iron_pickaxe.png}}. Compared to Multi-Modal Memory, {\mname} collects 1.8$\times$ more seeds~\raisebox{-0.3ex}{\includegraphics[width=0.3cm]{icon/seed.png}}, 1.5$\times$ more wood~\raisebox{-0.3ex}{\includegraphics[width=0.3cm]{icon/wood.png}}, 1.8$\times$ more dirt~\raisebox{-0.3ex}{\includegraphics[width=0.3cm]{icon/dirt.png}}, travels 1.3$\times$ further~\raisebox{-0.3ex}{\includegraphics[width=0.3cm]{icon/explore.png}}, and digs 1.1$\times$ deeper~\raisebox{-0.3ex}{\includegraphics[width=0.3cm]{icon/iron_pickaxe.png}}.
%
% %
% This shows that when executing single-step instructions, visual goal embeddings reconstructed from goal images generated by Dreamer provide more precise latent demonstrations of the desired behavior compared to direct text-to-visual space alignments~(\ie, STEVE-1~\cite{lifshitz2023steve}). Additionally, compared to Multi-Modal Memory, the visual goal embeddings generated by Dreamer are more in line with the current scene, improving the text-conditioned control over agent behavior.
%
This demonstrates that our CoI mechanism, which breaks down instructions into multiple stages and executes them step by step, leads to steadier instruction following compared to STEVE-1~\cite{lifshitz2023steve} which uses direct text instruction guidance. Unlike Multi-Modal Memory, which also features the CoI mechanism, our method generates future imaginations that closely resemble the current state at each stage, resulting in providing more precise visual prompts of the desired behavior, thus enhancing the stability of action generation.
%
% We also observe an interesting phenomenon: although Multi-Modal Memory outperforms the unconditional VPT(rl), it falls short in certain tasks compared to STEVE-1~\cite{lifshitz2023steve}, despite leveraging current observations.
% %
% Upon examining the generated videos and retrieval outcomes, we note that due to the vast diversity of open-world scenarios, the scenes executed in the videos retrieved by Multi-Modal Memory still show a little difference from the current situation, which in turn somewhat misleads the policy in predicting the agent's actions.

\begin{figure*}[t]
\centering
\includegraphics[width=1\linewidth]{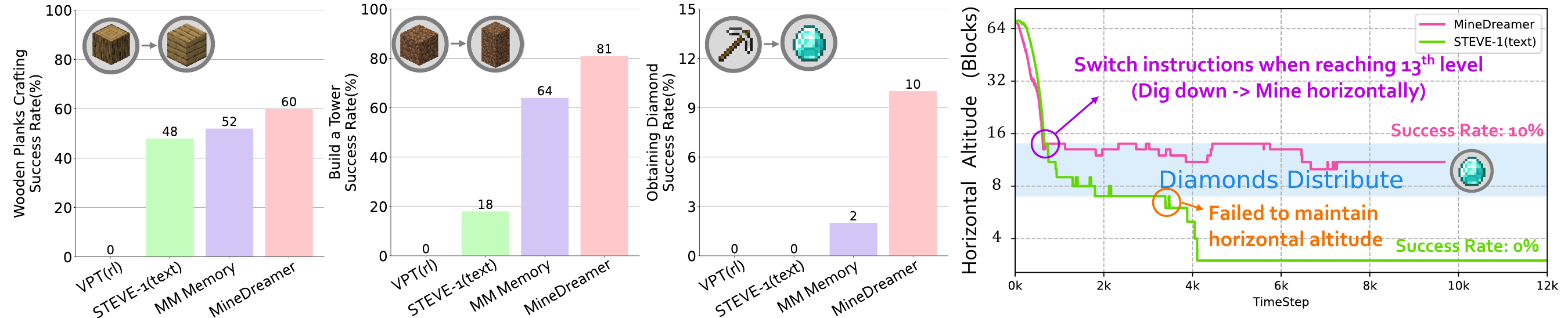}
\vspace{-5.5mm}
   \caption{\textbf{Performance on Command-Switching Evaluation. } \textbf{(Left)} {\mname} swiftly adapts to instructions and follows them steadily, achieving a higher success rate than the unconditional VPT~\cite{baker2022video}, the text-conditioned STEVE-1~\cite{lifshitz2023steve}, and the Multi-Modal Memory with CoI mechanism. \textbf{(Right)} {\mname} can dig down~\raisebox{-0.3ex}{\includegraphics[width=0.3cm]{icon/iron_pickaxe.png}} to a depth of 13 and steadily mine horizontally~\raisebox{-0.3ex}{\includegraphics[width=0.3cm]{icon/iron_pickaxe.png}} to obtain diamonds~\raisebox{-0.3ex}{\includegraphics[width=0.3cm]{icon/diamond.png}} with an average success rate of 10\%, while STEVE-1~\cite{lifshitz2023steve} struggles to maintain a consistent altitude.}
\vspace{-5.5mm}
\label{fig:switch_evaluation}
\end{figure*}

We also observe an interesting phenomenon: while Multi-Modal Memory, using the CoI mechanism and current observations, outperforms unconditional VPT~\cite{baker2022video}, it sometimes underperforms compared to STEVE-1~\cite{lifshitz2023steve}. Upon reviewing the recorded videos and the results of memory retrieval, we find that due to the vast diversity of open-world environments, the videos retrieved by Multi-Modal Memory still exhibit slight differences from the current state. This discrepancy misguides the PolicyNet in predicting agent actions, indicating that the CoI's effectiveness hinges on the relevancy and precision of future imaginations or visual prompts to the current state.
% \vspace{-2.5mm}

\vspace{-7.5mm}
\subsubsection{Command-Switching Evaluation for Long-Horizon Tasks.}
\label{subsub:Command-Switching Evaluation for Long-Horizon Tasks}
% \vspace{+1mm}
% \noindent{\textbf{Command-Switching Evaluation for Long-Horizon Tasks. }}
In this part, we explore agents' ability to solve long-horizon tasks that require executing multi-step instructions in sequence, including (1) collect wood~\raisebox{-0.3ex}{\includegraphics[width=0.3cm]{icon/wood.png}} and then craft planks~\raisebox{-0.3ex}{\includegraphics[width=0.3cm]{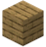}}, (2) gather dirt~\raisebox{-0.3ex}{\includegraphics[width=0.3cm]{icon/dirt.png}} and then build a tower~\raisebox{-0.3ex}{\includegraphics[width=0.3cm]{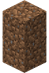}} and (3) dig down~\raisebox{-0.3ex}{\includegraphics[width=0.3cm]{icon/iron_pickaxe.png}} and then mine horizontally~\raisebox{-0.3ex}{\includegraphics[width=0.3cm]{icon/iron_pickaxe.png}} for diamonds~\raisebox{-0.3ex}{\includegraphics[width=0.3cm]{icon/diamond.png}}, each with 50 trials.
Tasks 1 and 2 limits 3,000 frames (\ie, 2.5 minutes of gameplay), with instructions changing at 1,500 and 2,000 frames. Task 3 limits 12,000 frames (\ie, 10 minutes of gameplay), switching instructions upon reaching the 13th floor, as diamonds~\raisebox{-0.3ex}{\includegraphics[width=0.3cm]{icon/diamond.png}} are commonly found between the 7th and 14th floors.
%
% sequentially

In \cref{fig:switch_evaluation}~(Left), {\mname} consistently surpasses VPT~\cite{baker2022video} and STEVE-1~\cite{lifshitz2023steve} in Command-Switching tasks. VPT 's~\cite {baker2022video} inability to follow instructions leads to a complete failure in executing sequential instructions, as evidenced by a 0\% success rate in the evaluation.
Although STEVE-1~\cite{lifshitz2023steve} occasionally completes Command-Switching tasks, it underperforms compared to {\mname}. For instance, in the \texttt{Obtain diamond}~\raisebox{-0.3ex}{\includegraphics[width=0.3cm]{icon/diamond.png}} task, STEVE-1's~\cite{lifshitz2023steve} success rate is 0\%, while Multi-Modal Memory's success rate is 2\%, notably lower than {\mname}'s 10\%.
As shown in \cref{fig:switch_evaluation}~(Right), we reconstruct an instance where two agents act in the same environment based on the simulator records. Initially, both {\mname} and STEVE-1~\cite{lifshitz2023steve} rapidly dig down~\raisebox{-0.3ex}{\includegraphics[width=0.3cm]{icon/iron_pickaxe.png}} to the target depth and then mine horizontally~\raisebox{-0.3ex}{\includegraphics[width=0.3cm]{icon/iron_pickaxe.png}} to obtain diamonds~\raisebox{-0.3ex}{\includegraphics[width=0.3cm]{icon/diamond.png}}.
Compared to STEVE-1~\cite{lifshitz2023steve}, {\mname} can consistently maintain the specified horizontal level over an extended period and successfully obtains diamonds~\raisebox{-0.3ex}{\includegraphics[width=0.3cm]{icon/diamond.png}} around the 10k steps in this instance.
While STEVE-1~\cite{lifshitz2023steve} manages to maintain its specified horizontal level for a long time, it ultimately fails to do so and becomes stuck in the bedrock layer~(\ie, the agent cannot break any block), resulting in a 0\% success rate.
This demonstrates that, even when instructions are switched rapidly, the CoI mechanism can still drive the agent to generate future goal imaginations that align with the current state. Visual prompts generated from these imaginations enable the agent to quickly adapt its actions to correspond with the new instructions while steadily following the instructions in action generation.

\begin{figure*}[htbp]
\centering
\includegraphics[width=\linewidth]{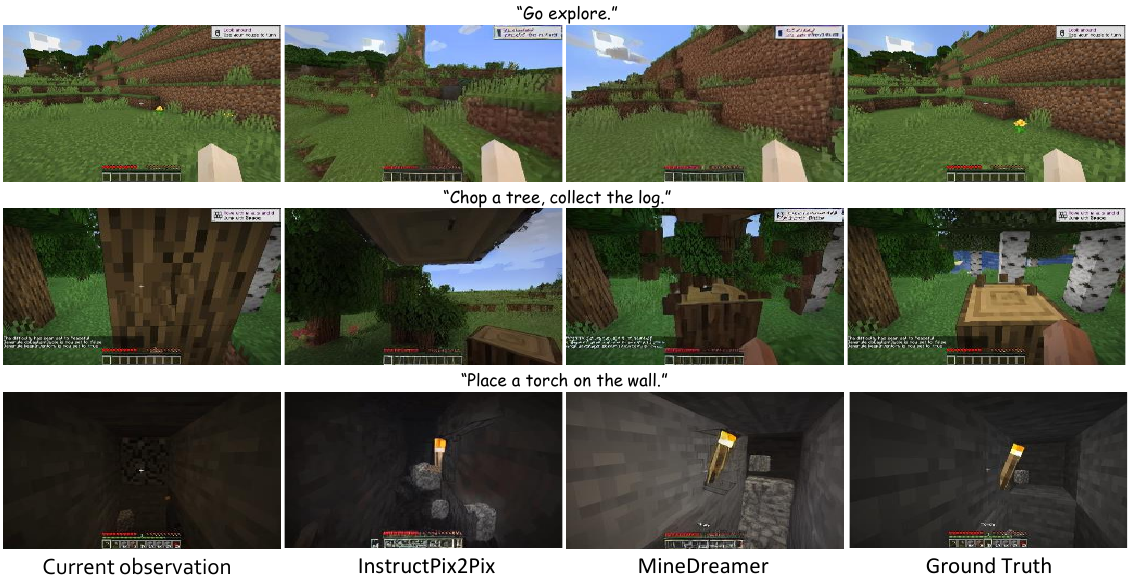}
\vspace{-6.5mm}
   \caption{Qualitative Comparison of Goal Imagination Generation. When compared to InstructPix2Pix~\cite{brooks2023instructpix2pix} that have undergone further fine-tuning on our Goal Drift Dataset, our approach demonstrates superior goal imagination capabilities in embodied scenarios. See Sec.~\ref{sub: Qualitative Results of Imaginator} for a more detailed analysis.}
\label{fig:evaluation}
\vspace{-7.5mm}
\end{figure*}

\vspace{-4mm}
\subsection{Qualitative Results of Imaginator}
\label{sub: Qualitative Results of Imaginator}

We compare Imaginator with the existing state-of-the-art instruction-based image editing model, namely InstructPix2Pix~\cite{brooks2023instructpix2pix}. Given this model has been trained on specific datasets, its performance would inevitably be suboptimal if directly applied to the Minecraft domain. To facilitate a fair comparison, we fine-tune InstructPix2Pix~\cite{brooks2023instructpix2pix} using the same training set employed by the Imaginator and assess the performance of the fine-tuned models in addressing tasks in Minecraft. Fig~\ref{fig:evaluation} shows qualitative results in the evaluation set, our methodology exhibits enhanced abilities in Goal Imagination Generation within intricate scenarios.

The first comparison shows that the Imaginator adeptly captures the agent's perspective shift as it advances, whereas InstructPix2Pix~\cite{brooks2023instructpix2pix} struggles to generate images in alignment with the provided instructions. In the second instance, the Imaginator specifically visualizes the region with felled trees~\raisebox{-0.4ex}{\includegraphics[width=0.3cm]{icon/tree.png}}, contrasting with InstructPix2Pix~\cite{brooks2023instructpix2pix}, which yields an image markedly divergent from the existing observation background. The third comparison highlights the Imaginator's ability to depict enhanced visibility following torch placement, in contrast to InstructPix2Pix~\cite{brooks2023instructpix2pix}, which merely adds torches without the associated increase in illumination.
These observations suggest that in scenarios requiring instruction reasoning and goal understanding, a simple CLIP~\cite{radford2019language} text encoder may struggle to guide the diffusion model to generate reasonable goal imagination. However, the MLLM can fully utilize its powerful reasoning ability, vast environmental knowledge, and intrinsic physical rules to correctly understand the goal and generate goal imagination. 
More visual results can be found in Supp.~\ref{sup:Visual} and Supp.~\ref{sup:Videos}.
% Goal Imagination Generation results when the agent executes tasks in Sup.~\todo{sup} further illustrate this point.

\vspace{-4mm}
\subsection{Discussion on Generalization}

In this part, we will explore the generalizability of {\mname}, as the agent's ability to generalize is key to its behavior in the open world where environments are complex and instructions vary widely. Since STEVE-1~\cite{lifshitz2023steve} has shown its prior ability to map text to visual prompts effectively, and our Prompt Generator is built upon it, we will now concentrate on the generalizability of our Imaginator and the entire agent.
%, using the same evaluation setup as in the Programmatic Evaluation.
%
At first, we exclude data related to the words `\texttt{Dirt}'~\raisebox{-0.3ex}{\includegraphics[width=0.3cm]{icon/dirt.png}} or `\texttt{Dig}'~\raisebox{-0.3ex}{\includegraphics[width=0.3cm]{icon/iron_pickaxe.png}} from the Goal Drift Dataset and retrain the model. Then, we observe the images generated in response to the instruction ``\texttt{Collect dirt}''~\raisebox{-0.3ex}{\includegraphics[width=0.3cm]{icon/dirt.png}} based on the current state and the quantity of dirt~\raisebox{-0.3ex}{\includegraphics[width=0.3cm]{icon/dirt.png}} collected by the agent.

\begin{figure*}[t]
\centering
\includegraphics[width=\linewidth]{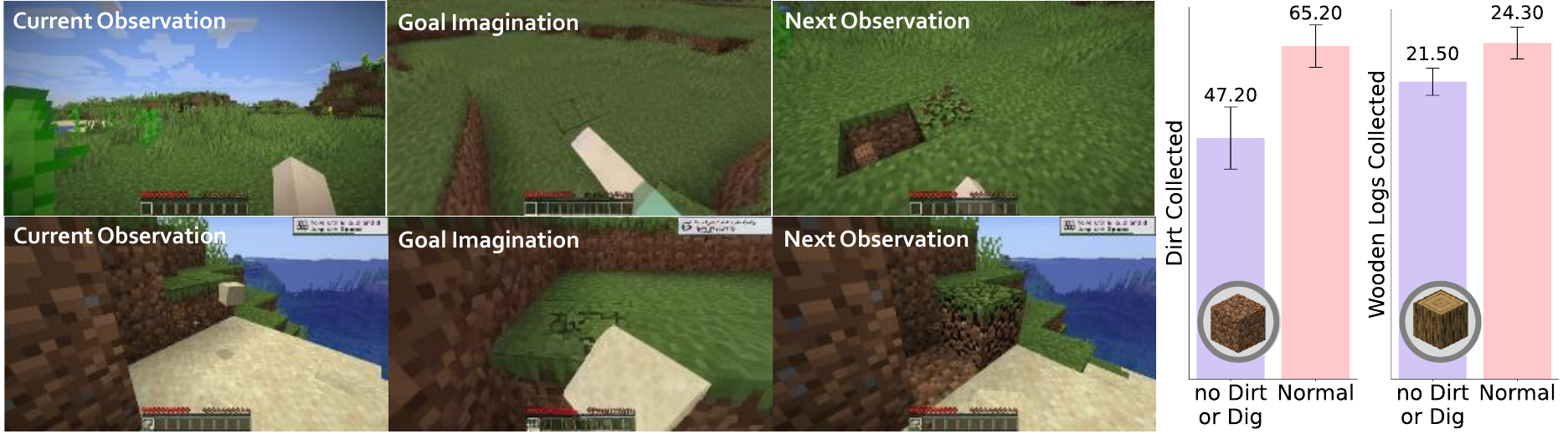}
\vspace{-7mm}
   \caption{\textbf{The Generalizability of {\mname}.} \textbf{(Left)} Despite excluding data involving `\texttt{Dirt}'~\raisebox{-0.3ex}{\includegraphics[width=0.3cm]{icon/dirt.png}} or `\texttt{Dig}'~\raisebox{-0.3ex}{\includegraphics[width=0.3cm]{icon/iron_pickaxe.png}} from Goal Drift Dataset and retraining, Imaginator can still generate relatively high-quality imaginations aligned with the instruction's concept. \textbf{(Right)} The retrained Imaginator remains operational with the CoI mechanism and can handle unseen instructions while largely preserving its previous performance.}
\vspace{-6mm}
\label{fig:General_result}
\end{figure*}
%%%%%%%%%%%%%%%%%%%%%%%%%%%%
\begin{table*}[b]
\vspace{-4mm}
\caption{We study the impact of dataset collection methods on agent performance. Values in parentheses represent 95\% confidence intervals.} 
\vspace{-4mm}
\centering

\scriptsize
\begin{tabular}{l||c||c|c|c}
\bottomrule[1pt]
 \multirow{2}{*}{Instruction}& Fixed Timestep      & Only Backward  &  Only Forward   &   Normal        \\
                             & Backward            &   Drift        &  Drift          &              \\
 \hline                         
``\texttt{Chop a tree}''~\raisebox{-0.4ex}{\includegraphics[width=0.3cm]{icon/tree.png}}     & 7.60(3.84, 11.36) & 10.10(2.82, 5.58) & 4.20(2.82, 5.58)       & \textbf{24.30(21.71, 26.89)} \\
``\texttt{Collect dirt}''~\raisebox{-0.3ex}{\includegraphics[width=0.3cm]{icon/dirt.png}}    & 38.60(21.97, 55.23) & 30.30(20.71, 39.89) & 18.10(6.74, 29.46) & \textbf{65.20(55.81, 74.59)} \\

\bottomrule[1pt]
\end{tabular}
\label{tab:dataset}
\end{table*}

As shown in \cref{fig:General_result}, we find that even after completely removing the concepts of `\texttt{Dirt}'~\raisebox{-0.3ex}{\includegraphics[width=0.3cm]{icon/dirt.png}} or `\texttt{Dig}'~\raisebox{-0.3ex}{\includegraphics[width=0.3cm]{icon/iron_pickaxe.png}}, Imaginator is still able to generate goal imaginations of relatively good quality aligned with the instruction's concept~(\ie, \textbf{agent points towards the dirt~\raisebox{-0.3ex}{\includegraphics[width=0.3cm]{icon/dirt.png}} and attempt to break it}), which can still guide the PolicyNet to follow instructions. 
The resulting collection of dirt~\raisebox{-0.3ex}{\includegraphics[width=0.3cm]{icon/dirt.png}} is about 70\% of the original amount, which shows that the Imaginator can respond to unseen novel instructions while largely maintaining its previous performance. 
We attribute this to three key factors: 
\textbf{(1)} The MLLM within Imaginator has the relevant environmental knowledge to map the text `\texttt{Dirt}'~\raisebox{-0.3ex}{\includegraphics[width=0.3cm]{icon/dirt.png}} to its corresponding element in Minecraft images, recognizing its visual counterpart; 
\textbf{(2)} training data for related tasks, such as ``\texttt{Collect seeds}''~\raisebox{-0.3ex}{\includegraphics[width=0.3cm]{icon/seed.png}}, enables the MLLM to comprehend the meaning of action `\texttt{Collect}' in Minecraft task; 
\textbf{(3)} The pre-trained Diffusion model can generalize to the Minecraft domain and generate goal imaginations by leveraging the MLLM's latent representations for understanding textual semantics mentioned above from the instructions.
\begin{table*}[htbp]
\vspace{-4mm}
\caption{We study the impact of the Chain-of-Imagination and diffusion model ability on agent performance. Values in parentheses represent 95\% confidence intervals.} 
\vspace{-4mm}
\centering

\scriptsize
\begin{tabular}{l||c||c|c||c}
\bottomrule[1pt]
 \multirow{2}{*}{Instruction} &  wo CoI & Random & Instruct- &   Normal        \\
                              &          & Noise  & Pix2Pix   &          \\
 \hline                         
``\texttt{Chop a tree}''~\raisebox{-0.4ex}{\includegraphics[width=0.3cm]{icon/tree.png}}     & 18.70(15.26, 22.14) & 2.70(0.85, 4.55) & 22.90(20.17, 25.63)       & \textbf{24.30(21.71, 26.89)} \\
``\texttt{Collect dirt}''~\raisebox{-0.3ex}{\includegraphics[width=0.3cm]{icon/dirt.png}}    & 53.50(36.93, 70.07) & 10.90(3.95, 17.85) & 59.50(54.00, 65.00) & \textbf{65.20(55.81, 74.59)} \\

\bottomrule[1pt]
\end{tabular}
\label{tab:coi_and_diffusion}
\vspace{-7mm}
\end{table*}

\vspace{-5mm}
\subsection{What Contributes to Performance}

\subsubsection{Dataset Collection Method.}
\label{subsub:Dataset Collection Method}
% \vspace{+1mm}
% \noindent{\textbf{Dataset Collection Method. }}
In \cref{tab:dataset}, we study the impact on agent performance by training with datasets of equal size collected using fixed Backward timesteps, only Backward Drift, only Forward Drift, and normal \textit{Goal Drift Dataset Collection}. 
Although data collected using the first three methods can enable the agent to follow instructions, the Imaginator is affected by Goal Illusion and Imagination Stagnation, which are discussed in \cref{subsub:Goal Drift Dataset Collection}. This results in the Imaginator's inability to envision the step-by-step process of completing the instruction and how to steadily complete the instruction multiple times.

\vspace{-4.5mm}
\subsubsection{Chain-of-Imagination.}
\label{subsub:Chain-of-Imagination}
% \vspace{+1mm}
% \noindent{\textbf{Chain-of-Imagination. }}
In \cref{tab:coi_and_diffusion}, we explore the effect of the CoI mechanism on agent performance, where ``wo-CoI'' denotes the scenario where the agent generates the goal imagination and visual prompt only at the beginning and remains unchanged thereafter. 
Compared to normal performance, ``wo-CoI'' achieves about 77\%. This is because the visual prompts generated at the beginning become less capable of providing precise demonstrations of the desired behavior in later stages, resulting in hindering the ability to guide the agent step by step more steadily.

\vspace{-4.5mm}
\subsubsection{Diffusion Model Ability.}
\label{subsub:Diffusion Model Ability}
% \vspace{+1mm}
% \noindent{\textbf{Diffusion Model Ability. }}
In \cref{tab:coi_and_diffusion}, we explore the impact of diffusion model ability on performance. Using ``random noise'' as a goal imagination results in vague visual prompts, which drastically reduce performance to merely 10\% of its original level.
The performance of InstructPix2Pix~\cite{brooks2023instructpix2pix} and our MLLM-enhanced diffusion model are comparable; however, by leveraging MLLM, our generated images adhere more closely to physical rules and environmental knowledge, as shown in \cref{fig:evaluation}. Additionally, as discussed in \cref{subsub:Programmatic Evaluation}, we discover that the CoI mechanism demands a certain quality of goal imagination, suggesting that the stronger the Imaginator, the better it can guide agents to follow instructions.

\vspace{+1mm}
\noindent More ablation studies can be found in Supp.~\ref{sup:Ablation}.

\vspace{-4mm}
\section{Conclusion and  Limitation}
\label{sec:Conclusion}
\vspace{-2mm}
In this paper, we introduce an innovative paradigm for enhancing the instruction-following ability of agents in simulated-world control.
We prove that by employing a Chain-of-Imagination mechanism to envision the step-by-step process of executing instructions, and translating imaginations into precise visual prompts tailored to the current state and instruction, 
can significantly help the foundation model follow instructions steadily in action generation. 
Our Agent, {\mname} in Minecraft, showcases its strong instruction-following ability. 
Furthermore, we show its potential as a high-level planner's downstream controller in the challenging ``\texttt{Obtain diamond}''~\raisebox{-0.3ex}{\includegraphics[width=0.3cm]{icon/diamond.png}} task. 
We believe this novel paradigm will inspire future research and generalize to other domains and open-world environments.

\vspace{+1mm}
\noindent \textbf{Limitation. } 
Firstly, generating high-quality imagination can take seconds, slowing down frequent-use scenarios. Speed enhancements via distillation~\cite{salimans2022progressive} and quantization~\cite{gholami2021survey} may mitigate this.
Secondly, the Imaginator may produce unrealistic hallucinations. Integrating world knowledge via methods such as RAG~\cite{lewis2020retrieval} or reducing MLLM hallucinations~\cite{MinerviniAHD2014} could mitigate this.

% \begin{abstract}
%   The abstract should concisely summarize the contents of the paper. 
%   While there is no fixed length restriction for the abstract, it is recommended to limit your abstract to approximately 150 words.
%   Please include keywords as in the example below. 
%   This is required for papers in LNCS proceedings.
%   \keywords{First keyword \and Second keyword \and Third keyword}
% \end{abstract}

% \section{Introduction}
% \label{sec:intro}

% \section{Related Work}
% \label{sec:related work}

% \clearpage  % TODO REVIEW/FINAL: This \clearpage needs to be removed from both review and camera-ready versions.

\bibliographystyle{splncs04}
\bibliography{main}

\appendix
\clearpage
\setcounter{page}{1}

{
\centering
\Large
\textbf{MineDreamer: Learning to Follow Instructions \\ via  Chain-of-Imagination  for \\Simulated-World Control}\\
\vspace{0.5em}Supplementary Material \\
\vspace{1.0em}
}

\noindent The supplementary document is organized as follows:
\begin{flushleft}

\begin{itemize}
    \item Sec.~\ref{sup:Environment}: Environment Setting, like observation and action space.
    \item Sec.~\ref{sup:Dataset}: Dataset composition and collection.
    \item Sec.~\ref{sup:Implementation}: Implementation Details, like training details.
    \item Sec.~\ref{sup:Experiment}: Experiment Details, like baseline and evaluation details.
    \item Sec.~\ref{sup:Ablation}: More Ablation Studies about {\mname}.
    \item Sec.~\ref{sup:Visual}: More Visual Results about Imagination in {\mname}.
    \item Sec.~\ref{sup:Videos}: Demo videos about {\mname}.
\end{itemize}
\end{flushleft}

\section{Minecraft Environment}
\label{sup:Environment}

Minecraft is a widely popular sandbox game that offers players the freedom to build and explore their worlds without limits, which also extends to AI agents as well.
Within the game, AI agents encounter situations that closely mirror real-world challenges, requiring them to make decisions and solve endless tasks in an open-world setting. 
Consequently, Minecraft is an ideal platform for AI evaluation and stands as an exemplary benchmark for AI testing, due to its vast freedom and open nature.
With the help of Minecraft, AI researchers can more easily simulate a wide variety of complex and dynamic environments and tasks, allowing them to conduct experiments that enhance the practical and applicable value of AI technologies.
We use MineRL~\cite{guss2019minerl} v1.0\footnote{\url{https://github.com/minerllabs/minerl/releases/tag/v1.0}}, which corresponds to Minecraft 1.16.5, as our simulation platform, ensuring an environment that is consistent with those used by VPT~\cite{baker2022video} and STEVE-1~\cite{lifshitz2023steve}.
In this version of MineRL~\cite{guss2019minerl}, a significant advancement over its predecessor~(\ie, MineRL v0.4.4), lies in the simulation environment. The environment now enables AI agents to interact in a manner entirely consistent with human players, eschewing primitive actions or script-based APIs. This approach presents a more complex and challenging scenario for AI research.
More specifically, AI agents experience the environment as humans do, solely through egocentric RGB images, devoid of any privileged in-game information. Additionally, their interactions with the environment are restricted to low-level keyboard and mouse actions.
Consequently, AI agents trained in this version of MineRL~\cite{guss2019minerl}~(\ie, MineRL v1.0) resemble embodied agents capable of performing various tasks in an open-world environment, demonstrating a higher degree of generalization.
Furthermore, the abundance of gaming videos available on the internet~(\eg, YouTube), provides AI researchers with the opportunity to harness these vast datasets for extensive pre-training, enabling the development of a foundation model in the sequential decision-making domain.

\subsection{Observation Space}
\label{supsub:Observation Space}
Our observation space aligns with that of human players, comprising simply the raw pixels from Minecraft. This includes the hotbar, health indicators, player hands, equipped items, and the game environment itself.
Specifically, the simulator produces RGB images with a resolution of 640x360. When the agent takes action within the environment, the simulator renders the player's first-person perspective with a field of view of 70 degrees. If the agent opens the inventory, the simulator will render the GUI interface along with the mouse cursor.
Notably, we do not employ privileged information such as voxels and lidar information available in MineDojo~\cite{fan2022minedojo}, which could be provided to the agent. During actual inference, the PolicyNet of {\mname} \textbf{\highlight{only accepts the raw RGB pixels observations as input}} that the agent can obtain from the environment and generates text-conditioned low-level action controls based on these observations, which are consistent with those used in VPT~\cite{baker2022video} and STEVE-1~\cite{lifshitz2023steve}.

\subsection{Action Space}
\label{supsub:Action Space}

\begin{table*}[t]
\caption{\textbf{Action Space utilized in the MineRL~\cite{guss2019minerl} simulator.} The action space primarily consists of 14 keyboard and mouse operations, with detailed descriptions sourced from the Minecraft wiki~(\url{https://minecraft.fandom.com/wiki/Controls}).} 
\centering
\small
\begin{tabular}{c|c|c|l}
\bottomrule[1pt]
Index & Action & Human Action & Description \\
\hline 
1 & Forward & key W & Move forward.\\
\hline 
2 & Back & key S & Move backward.\\
\hline
3 & Left & key A & Strafe left.\\
\hline
4 & Right & key D & Strafe right.\\
\hline
5 & Inventory & key E & Open or close GUI inventory. \\
\hline
6 & Drop & key Q & Drop a single item from the stack of items the player\\ 
  &      &       &  is currently holding. \\
\hline
7 & Jump & key Space & Jump. When in the water, it keeps the player afloat. \\
\hline
8 & Sneak & key left Shift & Move slowly in the current direction of movement. \\
\hline
9 & Sprint & key left Ctrl & Move fast in the current direction of movement. \\
\hline
10 & Attack & left Mouse  & Destroy blocks (hold down); Attack entity (click \\
   &        &      Button             & once); Pick up the stack of items or place the stack \\
   &        &                   &  of items in the GUI (click once) \\ 
\hline
11 & Use    & right mouse & Place the item being held or interact with the block \\
   &        &   Button                &  that the player is currently looking at. \\
\hline
12 & Hotbar.[1-9] & keys 1 - 9 & Switch the appropriate hotbar cell. \\
\hline
13 & Yaw & move  &Turning; aiming; camera movement.Ranging from\\
   &        &  Mouse X                 &  -180 to +180. \\
\hline
14 & Pitch & move  &Turning; aiming; camera movement.Ranging from\\
   &        &  Mouse Y                 &  -180 to +180. \\

% \arrayrulecolor{black}
\bottomrule[1pt]
\end{tabular}
\label{tab:actions}
\vspace{-4mm}
\end{table*}

As shown in \cref{tab:actions}, our action space encompasses a vast array of actions that are consistent with those of human players~(\ie, keyboard and mouse), including keypresses, mouse movements, and clicks. 
Excluding the ``\texttt{chat}'' action, which serves to initialize the agent with pre-defined conditions, more details can be found in Supp.~\ref{supsub:Rules}.
Keyboard presses and mouse clicks are binary functional actions~(\eg, ``\texttt{Forward}'', ``\texttt{Back}'', ``\texttt{Left}'', ``\texttt{Right}'' and \etc). Beyond these binary input options, our action space also has mouse 
 cursor movements. While the GUI is closed (\ie, activated by pressing ``E'' for the GUI inventory) and remains inactive, the mouse's horizontal and vertical movements direct the agent's yaw and pitch. Conversely, with GUI open, the same movements are re-purposed to navigate the cursor across the display.

It is noteworthy that we have not employed structured APIs such as ``\texttt{craft}'' and ``\texttt{smelt}'' as seen in MineDojo~\cite{fan2022minedojo}, which replace the need for precise mouse movements that are necessary for interacting with the inventory for certain tasks, effectively turning these operations into GUI functional binary actions. 
During actual inference, our {\mname}'s PolicyNet \textbf{\highlight{only outputs keyboard and mouse actions}} to dictate the agent's movements, aligning these actions with those utilized in VPT~\cite{baker2022video} and STEVE-1~\cite{lifshitz2023steve}.

\subsection{Environment Settings and Rules}
\label{supsub:Rules}

In our experiments, the agent's \textit{initial position} at the start of the game, as well as the \textit{seed} used to generate the environment, are completely random. This introduces an element of unpredictability and variety into the experimental setup, ensuring that the agent will encounter a wide range of scenarios and challenges.
To better evaluate the agent's ability to follow textual instructions for action prediction and its ability to rapidly adapt its behavior based on instructions, we have modified MineRL~\cite{guss2019minerl} to enable ``\texttt{chat}'' action operations. This allows for the swift initialization of the agent with predefined conditions through instructions.
Specifically, for Programmatic Evaluation, we ensure that each experiment for all agents is conducted with the same seed and within the biome most conducive to completing the current instruction; across multiple experiments, different seeds are used.
For Command-Switching Evaluation for Long-Horizon Tasks, all agents are placed in the same seed and biome optimal for the current instruction as well. In addition, the following rules are applied as aids:
\begin{itemize}
    \small
    \item \texttt{/difficulty peaceful}: Set the difficulty of the environment to peaceful mode.
    \item \texttt{/gamerule doDaylightCycle false}: Set the environment to daytime forever.
    \item \texttt{/gamerule keep inventory true}: Set agent to not drop items upon death.
\end{itemize}
Specifically, for the task of ``Obtain diamonds''~\raisebox{-0.3ex}{\includegraphics[width=0.3cm]{icon/diamond.png}}, we add two additional rules on top of the aforementioned ones as assistance:
\begin{itemize}
    \small
    \item \texttt{/effect give @a night\_vision 99999 250 true}: Help the agent see more clearly in extremely dark environments (\eg, at night or underground).
    \item \texttt{/give @p minecraft:diamond\_pickaxe}: Provided the agent with a diamond pickaxe~\raisebox{-0.3ex}{\includegraphics[width=0.3cm]{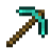}}, enabling it to break almost all blocks and mine all ores within Minecraft.
\end{itemize}
For details regarding the most suitable biome used in the experiments, please check Supp.~\ref{supsub:Programmatic} and Supp.~\ref{supsub:Command-Switching}.

\section{Dataset Details}
\label{sup:Dataset}

\subsection{OpenAI Contractor Gameplay Dataset}
\label{supsub:OpenAI Contractor Gameplay Dataset}
All our raw data are based on the contractor dataset\footnote{\url{https://github.com/openai/Video-Pre-Training}}, which consists of offline trajectory data in Minecraft used for training VPT~\cite{baker2022video}.
This dataset is created by hiring human contractors to play Minecraft and complete predetermined tasks, and it includes video~(\ie, image sequences), along with corresponding action sequences and metadata.
OpenAI releases six subsets of contractor data: 6.x, 7.x, 8.x, 9.x, 10.x, and the MineRL BASALT 2022 dataset. Our Goal Drift Dataset ultimately selects three of these subsets as our raw data, including 8.x (house building from scratch), 10.x (~\raisebox{-0.3ex}{\includegraphics[width=0.3cm]{icon/diamond_pickaxe.png}}), and the \texttt{FindCave} dataset from the MineRL BASALT 2022 dataset.
For each video, there is an associated metadata file that not only records the contractor's actions for every frame but also documents events triggered by the contractor within the simulator; the specific events are detailed in \cref{tab:event}.

\begin{table*}[htbp]
\caption{\textbf{The detailed event name and description in MineRL~\cite{guss2019minerl} simulator.} The simulator records the names of events that occur as well as related information, including quantities. We can use these events to collect a large amount of data for completing event-related instruction tasks with clarity.} 
\vspace{-4mm}
\centering
\scriptsize
\begin{tabular}{l|l}
\bottomrule[1pt]
Event Name & Description\\
\hline 
\texttt{mine\_block} & The moment the agent breaks a block, the type of block is recorded. \\
\texttt{craft\_item} & The moment the agent crafts items, the type and number of items are recorded. \\
\texttt{use\_item} & The moment the agent uses or places items, the type of item is recorded. \\
\texttt{kill\_entity} & The moment the agent kills an entity, the type of entity is recorded. \\
\texttt{break\_item} & The moment the tool of the agent is broken, the type of tool is recorded. \\
\texttt{pick\_up} & The moment the agent picks up items, the type and number of items are recorded. \\

% \arrayrulecolor{black}
\bottomrule[1pt]
\end{tabular}
\label{tab:event}
\vspace{-4mm}
\end{table*}

\subsection{Event Selection}
\label{supsub:Event Selection}
In constructing our dataset, we opt to \textbf{\highlight{select events directly from the MineRL simulator and supplement them with manually annotated events}}. 
Specifically, to train the Imaginator within the constraints of limited resources, we focus on the following types of events: ``\texttt{mine\_block}'', ``\texttt{craft\_item}'', ``\texttt{use\_item}'', ``\texttt{kill\_entity}'' and a manually defined event named ``\texttt{easy\_action}''.
Details of the specific items selected for each event can be found in \cref{tab:event selection}.
The simulator's built-in events have a clearly defined completion time \( t^* \), while manually annotated events are marked with a manually labeled completion time.

\begin{table*}[htbp]
\caption{\textbf{Details of the specific items selected for each event.} We select four built-in events from the simulator, along with a manually defined event called ``\texttt{easy\_action}''. The built-in events have a clearly defined completion moment, while the collection of the ``\texttt{easy\_action}'' event is manually annotated.} 
\centering
\small
\begin{tabular}{|c|c|c|c|c|c|}
\bottomrule[1pt]
Event Name                      &  \texttt{mine\_block}    &  \texttt{craft\_item} & \texttt{use\_item} & \texttt{kill\_entity}  &    \texttt{easy\_action}          \\
\hline
\multirow{11}{*}{Detail items}  & Wooden Log               &   wooden planks       & torch              & sheep                 & Go explore                 \\
                                & Grass                    &                       &                    &                       & Dig down                      \\
                                & Dirt                     &                       &                    &                       & Look at the sky                 \\
                                & Grass Block              &                       &                    &                       & Go Swimming                      \\
                                & Sand                     &                       &                    &                       & Stay underwater                  \\
                                & Snow                     &                       &                    &                       & Build a tower                   \\
                                & Stone                    &                       &                    &                       & Mine horizontally              \\
                                & Coal Ore                 &                       &                    &                       &    \\ 
                                & Iron Ore                 &                       &                    &                       &    \\
                                & Redstone Ore             &                       &                    &                       &    \\ 
                                & Diamond Ore              &                       &                    &                       &    \\

% \arrayrulecolor{black}
\bottomrule[1pt]
\end{tabular}
\label{tab:event selection}
\vspace{-4mm}
\end{table*}

\subsection{Dataset Collection}
\label{supsub:Dataset Collection}
After obtaining the completion times \( t^* \) for all events, we employ \texttt{gpt-4-turbo}~\cite{gpt4v} to generate corresponding event-related instructions.
Specifically, we provide \texttt{gpt-4-turbo}~\cite{gpt4v} with the event's name, description, and detailed items, and prompt it to generate multiple distinct simple instructions. These instructions include specific actions, while others mention the items to be obtained upon completing the action. For instance, for ``\texttt{Grass}'' in the ``\texttt{mine\_block}'' event, \texttt{gpt-4-turbo}~\cite{gpt4v} would generate instructions like ``break grass'', ``break tall grass'', ``gather seeds'', and ``collect seeds''.
After gathering instructions for all events, we apply the \textit{Goal Drift Collection} method described in \cref{subsub:Goal Drift Dataset Collection} of the main paper to conduct backward and forward drift on the completion times \( t^* \) of event-related instructions. For each pair (current observation, goal imagination), there are many instructions created by \texttt{gpt-4-turbo}~\cite{gpt4v} to describe that event. This process results in a substantial collection of triplets (current observation, goal imagination, instruction), which serve as training data for the Imaginator, forming what we call the Goal Drift Dataset.
The final Goal Drift Dataset contains approximately 500,000 triplets (current observation, goal imagination, instruction), with about 400,000 of these triplets derived from events built into the simulator.
We follow the method used in STEVE-1~\cite{lifshitz2023steve} for training the CVAE~\cite{sohn2015learning, kingma2013auto} and collect a subset of approximately 10,000 quadruplets from the Goal Drift Dataset for the events we need to test subsequently. This subset consists of quadruplets where the current observation, goal imagination, and instruction are consistent as conditions with the Goal Drift Dataset. Additionally, there is a visual prompt embedding that serves as ground truth. This embedding is derived from a video composed of the goal imagination and the preceding 16 frames, processed through the MineCLIP~\cite{fan2022minedojo} video encoder.

\section{Implementation Details}
\label{sup:Implementation}
\subsection{Imaginator}
\label{supsec:Imaginator}
The training process of Imaginator is divided into three main stages. In the first stage, the MLLM is aligned with the CLIP~\cite{radford2021learning} text encoder using the QFormer~\cite{li2023blip}. In the second stage, we apply InstructPix2Pix~\cite{brooks2023instructpix2pix} to warm up the weights for the diffusion model in Minecraft. In the third stage, we optimize Imaginator in an end-to-end manner. To be specific, the weights of LLaVA~\cite{liu2023visual} are frozen and LoRA~\cite{hu2021lora} is added for efficient fine-tuning. For the diffusion model, we directly use the weights pre-trained in the second stage as the initial weights in Imaginator. 

For the Large Language Model with visual input (e.g., LLaVA~\cite{liu2023visual}), we choose LLaVA-1.1-7b~\cite{liu2023visual} as the base model. During training, the weights of LLaVA are frozen and we add LoRA for efficient fine-tuning. We expand the original LLM vocabulary with $32$ new tokens. The QFormer is composed of $6$ transformer~\cite{vaswani2017attention} layers and $77$ learnable query tokens. We use the AdamW optimizer~\cite{loshchilov2017decoupled} in all three stages. 
In the initial stage of training, we configure the learning rate and weight decay parameters at 2e-4 and 0, respectively. The training targets for this stage encompass a dual-objective framework, comprising the Mean Squared Error~(MSE) loss between the outputs of LLaVA~\cite{liu2023visual} and the CLIP~\cite{radford2021learning} text encoder, alongside the language model loss. Both losses are assigned equal weights of 1.
The training setting in the second is the same as InstructPix2Pix~\cite{brooks2023instructpix2pix}.
In the final stage, the settings for the learning rate, weight decay, and warm-up ratio are adjusted to 1e-5, 0, and 0.001, respectively. During this phase, the loss function is diffusion loss. 

\begin{table*}[htbp]
\caption{\textbf{The Hyperparameters of Imaginator.}} 
\centering
\small
\begin{tabular}{|l|c|}
\bottomrule[1pt]
\text { Hyperparameter Name } & \text { Value } \\
\hline \text { base\_model } & \text { LLaVA~\cite{liu2023visual} } \\
\hline \text { input\_image\_size } & \text { 256 $\times$ 256 } \\
\hline \text { expand\_vocabulary\_num } & 32 \\
\hline \text { transformer\_layers\_num } & 6 \\
\hline \text { QFormer\_learnable\_query\_num} & 77 \\
\hline \text { optimizer } & AdamW~\cite{loshchilov2017decoupled} \\
\hline \text { learning\_rate\_initial\_stage  } & 2e-4 \\
\hline \text { weight\_decay\_initial\_stage  } & 0 \\
\hline \text { learning\_rate\_final\_stage } & 1e-5 \\
\hline \text { weight\_decay\_final\_stage } & 0 \\
\hline \text { warm-up\_ratio\_final\_stage } & 0.001 \\
\hline \text { n\_iterations\_initial\_stage } & 5000 \\
\hline \text { n\_iterations\_final\_stage } & 10000 \\
% \arrayrulecolor{black}
\bottomrule[1pt]
\end{tabular}
\label{tab:CVAE Hyperparameters}
\vspace{-4mm}
\end{table*}

\subsection{Prompt Generator}
\label{supsec:Prompt Generator}
Our Prompt Generator is mainly a conditional variational autoencoder (CVAE)~\cite{sohn2015learning, kingma2013auto} with a Gaussian prior and a Gaussian posterior similar to STEVE-1~\cite{lifshitz2023steve}. 
Both the encoder and decoder of CVAE~\cite{sohn2015learning, kingma2013auto} are parameterized as three-layer MLPs with 512 hidden units and layer normalization.
It encodes the current observations, goal imaginations, and instructions then reconstructs a latent visual embedding, and uses a linear layer to project this embedding into the visual input space of our PolicyNet as the final visual prompt.
It is noteworthy that instead of using raw pixel images and natural language instructions directly as conditions to generate pixel-level videos depicting the execution of an instruction from the current observation to the imagined target, we opt to perform reconstruction within the visual space of MineCLIP~\cite{fan2022minedojo}, where MineCLIP~\cite{fan2022minedojo} is a pre-trained CLIP model that employs a contrastive objective on pairs of Minecraft videos and associated transcripts from the web.
Specifically, the process of generating prompts by the Prompt Generator mainly involves three steps.
First, we stack the current observation and the goal imagination 16 times each to create two static 16-frame videos. These are then processed through MineCLIP~\cite{fan2022minedojo}'s video encoder to obtain two visual embeddings. Concurrently, the instruction is encoded into a text embedding using MineCLIP~\cite{fan2022minedojo}'s text encoder. This ensures that all embeddings are encoded within the MineCLIP~\cite{fan2022minedojo} space.
We then train a CVAE~\cite{sohn2015learning, kingma2013auto} using the ELBO loss, which reconstructs a latent visual embedding from the previous three embeddings. This representation is a video embedding that describes the process within the MineCLIP~\cite{fan2022minedojo} visual space.
This representation is a video embedding that captures the process within the MineCLIP~\cite{fan2022minedojo} visual space. The ground truth for this is mentioned in Supp.~\ref{supsub:Dataset Collection} and is derived from the goal imagination and the preceding 16 frames, which have been processed through the MineCLIP~\cite{fan2022minedojo} video encoder.
In the end, we use a linear layer to project the latent visual embedding into the visual input space of the PolicyNet as the final visual prompt.
For each event to be evaluated subsequently, we train a CVAE~\cite{sohn2015learning, kingma2013auto} on the dataset, specifically for 150 epochs with early stopping on a small validation set.
Notably, the parameters of the MineCLIP~\cite{fan2022minedojo} within Prompt Generator remain unchanged, as do the parameters of the linear layer that maps MineCLIP~\cite{fan2022minedojo}'s visual space to the visual input space of PolicyNet, whose parameters come from STEVE-1~\cite{lifshitz2023steve}. The hyperparameters used during the training are listed in the following \cref{tab:CVAE Hyperparameters}.

\begin{table*}[htbp]
\caption{\textbf{The Hyperparameters of CVAE~\cite{sohn2015learning, kingma2013auto} within Prompt Generator.}} 
\centering
\small
\begin{tabular}{|l|c|}
\bottomrule[1pt]
\text { Hyperparameter Name } & \text { Value } \\
\hline \text { architecture } & \text { MLP } \\
\hline \text { visual\_prompt\_dim } & 512 \\
\hline \text { text\_dim } & 512 \\
\hline \text { current\_img\_dim } & 512 \\
\hline \text { goal\_img\_dim } & 512 \\
\hline \text { hidden\_layers } & 3 \\
\hline \text { batch\_size } & 256 \\
\hline \text { learning\_rate } & 1e-4 \\
\hline ~$\beta$ & 0.001 \\
\hline \text { n\_epochs } & 150 \\
% \arrayrulecolor{black}
\bottomrule[1pt]
\end{tabular}
\label{tab:CVAE Hyperparameters}
\vspace{-4mm}
\end{table*}

\section{Experiment Details}
\label{sup:Experiment}

In this section, we first detail the three baselines we select. We then separately present the Programmatic Evaluation details and the Command-Switching Evaluation for Long-Horizon Tasks details.

\subsection{Baseline Datails}
\label{supsub:Baseline Datails}

\textbf{Video Pretraining~(VPT)}~\cite{baker2022video} is the first foundation model in the Minecraft domain, pre-trained on 70k hours of gameplay by Baker et al.~\cite{baker2022video}. Its architecture primarily consists of two parts: ImpalaCNN and TransformerXL~\cite{dai2019transformer}. VPT~\cite{baker2022video} has three variants: VPT(fd), VPT(bc), and VPT(rl), representing the vanilla foundation model, the behavior cloning fine-tuned model, and the RL fine-tuned model, respectively.
Specifically, they initially pre-trained on a large corpus of YouTube videos using a behavior cloning algorithm to obtain VPT(fd), which is capable of free exploration within the environment. This model gains a fundamental understanding of the environment and acquires some environmental knowledge.
To enhance the agent's capability in completing early-game tasks~(\eg, ``Collect wood''~\raisebox{-0.3ex}{\includegraphics[width=0.3cm]{icon/wood.png}} and ``Craft wooden planks''~\raisebox{-0.3ex}{\includegraphics[width=0.3cm]{icon/plank.png}}, they collect an ``Early-Game'' video dataset and fine-tune the VPT(fd) to obtain VPT(bc). This model performs well in early-game tasks but struggles with long-horizon tasks, such as obtaining diamonds~\raisebox{-0.3ex}{\includegraphics[width=0.3cm]{icon/diamond.png}}.
Building on VPT(bc), they employ online reinforcement learning with carefully designed rewards to fine-tune the model, enabling it to complete the task of obtaining diamonds~\raisebox{-0.3ex}{\includegraphics[width=0.3cm]{icon/diamond.png}} from scratch, ultimately resulting in the creation of VPT(rl).
Hence, it is noteworthy that \textbf{\highlight{all three variants of VPT~\cite{baker2022video} are unable to follow instructions}}; they must first be fine-tuned on downstream tasks before they can be completed. Despite their extensive environmental knowledge, this knowledge cannot be unlocked by instruction-following capabilities.
In our experiments, we use VPT(rl) because it initially seeks out trees~\raisebox{-0.4ex}{\includegraphics[width=0.3cm]{icon/tree.png}} and gathers wood~\raisebox{-0.3ex}{\includegraphics[width=0.3cm]{icon/wood.png}}, a critical step in the pathway to obtaining diamonds~\raisebox{-0.3ex}{\includegraphics[width=0.3cm]{icon/diamond.png}}. When set in the appropriate biome, VPT(rl) explores further~\raisebox{-0.3ex}{\includegraphics[width=0.3cm]{icon/explore.png}} and collects more wood~\raisebox{-0.3ex}{\includegraphics[width=0.3cm]{icon/wood.png}} compared to VPT(fd) and VPT(bc).

\vspace{+2mm}
\noindent\textbf{STEVE-1}~\cite{lifshitz2023steve} is a Minecraft agent that can follow both textual and visual instructions, built upon MineCLIP~\cite{fan2022minedojo} and VPT~\cite{baker2022video}. Drawing from the paradigms of instruction tuning in large language models and multimodal large language models, it successfully unlocks the instruction-following abilities of the foundation model~(\ie, VPT~\cite{baker2022video}) in the domain of decision-making.
STEVE-1~\cite{lifshitz2023steve} comes in two variants, STEVE-1(visual) and STEVE-1(text). The training process is divided into two steps. The first step involves training a policy conditioned on future video as visual instructions using the packed hindsight relabeling method. Specifically, they utilize the OpenAI Contractor Gameplay Dataset to fine-tune VPT(rl) to follow visual instructions, resulting in STEVE-1(visual).
The second step is to train a model that can map text instructions to visual instructions. Inspired by UnCLIP, they trained a Conditional Variational Autoencoder (CVAE)~\cite{sohn2015learning, kingma2013auto} on a dataset of video-text pairs they collected, thus obtaining STEVE-1(text) which can follow text instructions.
It is important to note that the visual or \textbf{\highlight{textual instruction variants of STEVE-1~\cite{lifshitz2023steve} do not consider the current observation and remain unchanged throughout the task, serving as an initial guide without adapting to environmental changes}}.

\vspace{+2mm}
\noindent\textbf{Multi-Modal Memory} serves as a substitute for the Imaginator and Prompt Generator in the \mname framework, essentially functioning by supplying PolicyNet with video prompts that best align with the current observations and textual instructions, similar to the approach of STEVE-1 (visual).
We construct a multi-modal memory comprised of numerous video-text pairs. This memory is specifically built upon the triplets (current observation, goal imagination, instruction) from the Goal Drift Dataset. By tracing back 16 frames from the timestamp of the goal imagination, we create a 16-frame video segment, resulting in a revised triplet format: (current observation, goal imagination video, instruction). Each event, whether from the MineRL~\cite{guss2019minerl} environment or manually defined, contains 1,000 pairs.
The retrieval process is as follows: First, we encode the current instruction and all instructions in the multi-modal memory using the OpenCLIP~\cite{Radford2021LearningTV} text encoder to obtain embeddings. We then compare these embeddings using cosine similarity. Next, within the memory corresponding to the text instruction with the highest similarity, we find the match where the current observation and the memory's observation, once encoded through the OpenCLIP~\cite{Radford2021LearningTV} Image encoder, have the highest cosine similarity in their embeddings. Finally, the video from the final retrieval result is then encoded using the MineCLIP~\cite{fan2022minedojo} video encoder, and the resulting visual embedding is used as the final visual prompt.
Therefore, \textbf{\highlight{Multi-Modal Memory leverages the current observation and also utilizes the Chain-of-Imagination~(CoI) mechanism}}.

\subsection{Programmatic Evaluation Datails}
\label{supsub:Programmatic}

In this part, we will elaborate on the selection of experimental tasks for Programmatic Evaluation, the methodology for calculating evaluation metrics, and the specific details of the experimental setup.

For the Programmatic Evaluation, we evaluate the agents on five \textbf{single-step} instruction tasks derived from the \textit{early-game evaluation suite} proposed in Table 3 of the STEVE-1~\cite{lifshitz2023steve} appendix. The purpose of this evaluation is to quantitatively measure an agent's ability to follow instructions with minimal human intervention.
Specifically, we calculate the programmatic evaluation metrics by monitoring the state of the MineRL~\cite{guss2019minerl} environment during each evaluation episode. Consistent with VPT~\cite{baker2022video} and STEVE-1~\cite{lifshitz2023steve}, we compute multiple programmatic metrics, including travel distance, dig depth, and early-game item collection. The calculation is as follows:
\begin{enumerate}
\small
    \item \texttt{Travel Distance (Blocks)}: The agent's maximum horizontal displacement, in the X-Z plane, is measured from the initial spawn point.
    \item \texttt{Dig Depth (Blocks)}: The agent's maximum vertical (Y-axis) displacement is measured from its initial spawn point.
    \item \texttt{Early-Game Inventory Counts}:  The maximum number of log~\raisebox{-0.3ex}{\includegraphics[width=0.3cm]{icon/wood.png}}, seed~\raisebox{-0.3ex}{\includegraphics[width=0.3cm]{icon/seed.png}}, and dirt~\raisebox{-0.3ex}{\includegraphics[width=0.3cm]{icon/dirt.png}} items seen in the agent’s inventory during the episode.
\end{enumerate}

We test all agents on these five \textbf{single-step} instruction tasks, with each task running 10 episodes of 3000 timesteps~(\ie, 2.5 minutes of gameplay). Each episode used a unique environmental seed, yet all agents were tested under the same seed for consistency.
It is important to note a key difference in our experimental setup compared to STEVE-1~\cite{lifshitz2023steve}: \textbf{\highlight{for each task, we initialize the agents in the biome most conducive to task completion}} to enhance the reliability of our evaluation metrics. For instance, in the ``\texttt{Chop a tree}''~\raisebox{-0.4ex}{\includegraphics[width=0.3cm]{icon/tree.png}} task, all agents are spawned in a forest biome, rather than a plain, to avoid the added randomness of searching for trees~\raisebox{-0.4ex}{\includegraphics[width=0.3cm]{icon/tree.png}} before chopping them. 
Due to a limited computational budget, we do not generate goal imaginations for every frame within an episode. In MineRL~\cite{guss2019minerl}, an agent can perform only one mouse or keyboard action per frame, and for tasks such as breaking a block of dirt~\raisebox{-0.3ex}{\includegraphics[width=0.3cm]{icon/dirt.png}}, it requires approximately 25 frames of consistently holding down the left mouse button. Therefore, \textbf{\highlight{we decide to imagine a goal imagination and translate it to a visual prompt every 25 frames}} ultimately, which then guides the action generation for the following 25 frames~(\ie, the visual prompt $p_t$ will not change for the next 25 frames). 
This interval is chosen because, aside from the ``\texttt{Chop a tree}''~\raisebox{-0.4ex}{\includegraphics[width=0.3cm]{icon/tree.png}} task, the other four tasks can be achieved within 25 frames~(\ie, just over 1 second of gameplay), thereby necessitating a new round of imagination to guide subsequent actions.
The detailed settings for the Programmatic Evaluation can be found in \cref{tab:Programmatic Evaluation Datails}.

\begin{table*}[htbp]
\caption{\textbf{The detailed settings for the Programmatic Evaluation.}} 
\centering
\small
\begin{tabular}{c|c|c|c|c|c}
\bottomrule[1pt]
Id & Text Instruction & Biome & Time Limit & Imagination Interval & Metric  \\
\hline 
1 & \texttt{go explore}~\raisebox{-0.3ex}{\includegraphics[width=0.3cm]{icon/explore.png}} & Plains & \multirow{5}{*}{3000 Frames} & \multirow{5}{*}{25 Frames} & Travel Distance (Blocks) \\

2 & \texttt{collect seeds}~\raisebox{-0.3ex}{\includegraphics[width=0.3cm]{icon/seed.png}} & Plains &  &  & Seeds Collected \\

3 & \texttt{chop a tree}~\raisebox{-0.4ex}{\includegraphics[width=0.3cm]{icon/tree.png}} & Forest &  &  & Wooden Logs Collected \\

4 & \texttt{collect dirt}~\raisebox{-0.3ex}{\includegraphics[width=0.3cm]{icon/dirt.png}} & Plains &  &  & Dirt Collected \\

5 & \texttt{dig down}~\raisebox{-0.3ex}{\includegraphics[width=0.3cm]{icon/iron_pickaxe.png}} & Plains &  &  & Dig Depth (Blocks) \\

% \arrayrulecolor{black}
\bottomrule[1pt]
\end{tabular}
\label{tab:Programmatic Evaluation Datails}
\vspace{-4mm}
\end{table*}

\subsection{Command-Switching Evaluation Datails}
\label{supsub:Command-Switching}

In this part, we will also detail the selection of experimental tasks for Command-Switching Evaluation for Long-Horizon Tasks, the calculation methods for evaluation metrics, and the specific details of the experimental setup.

The Command-Switching Evaluation for Long-Horizon Tasks comprises three \textbf{multi-step} instructions tasks sourced from the \textit{early-game evaluation suite} of STEVE-1~\cite{lifshitz2023steve}, except the ``\texttt{Obtain diamonds}''~\raisebox{-0.3ex}{\includegraphics[width=0.3cm]{icon/diamond.png}} task which originates from GROOT~\cite{cai2023groot}, designed to steadily follow video instructions. These tasks aim to evaluate an agent's ability to swiftly adapt to new instructions following an instruction switch, a critical capability for a downstream controller operating under an LLM-based high-level planner.
We employ success rate as the performance metric, also by monitoring the MineRL~\cite{guss2019minerl} environment state throughout each evaluation episode. The criteria for determining success across the three different tasks are as follows:
\begin{enumerate}
\small
    \item \texttt{collect wood~\raisebox{-0.3ex}{\includegraphics[width=0.3cm]{icon/wood.png}} and then craft planks~\raisebox{-0.3ex}{\includegraphics[width=0.3cm]{icon/plank.png}}}: Success is defined as successfully crafting at least one wooden log~\raisebox{-0.3ex}{\includegraphics[width=0.3cm]{icon/wood.png}} into four wooden planks~\raisebox{-0.3ex}{\includegraphics[width=0.3cm]{icon/plank.png}} within the given time frame.
    \item \texttt{gather dirt~\raisebox{-0.3ex}{\includegraphics[width=0.3cm]{icon/dirt.png}} and then build a tower~\raisebox{-0.3ex}{\includegraphics[width=0.3cm]{icon/tower.png}}}: Success is defined as successfully building a tower~\raisebox{-0.3ex}{\includegraphics[width=0.3cm]{icon/tower.png}} with a height of at least 7 blocks within the given time frame.
    \item \texttt{dig down~\raisebox{-0.3ex}{\includegraphics[width=0.3cm]{icon/iron_pickaxe.png}} and then mine horizontally~\raisebox{-0.3ex}{\includegraphics[width=0.3cm]{icon/iron_pickaxe.png}}}: Success is obtaining at least one diamond~\raisebox{-0.3ex}{\includegraphics[width=0.3cm]{icon/diamond.png}} within the given time frame.
\end{enumerate}

For these three \textbf{multi-step} instructions tasks, we run 50 episodes of testing per task. The time limit for the first two tasks is set at 3000 frames~(\ie, 2.5 minutes of gameplay), consistent with STEVE-1~\cite{lifshitz2023steve}, while the final task has an episode time limit of 12,000 frames~(\ie, 10 minutes of gameplay), aligning with what is mentioned in the main paper of GROOT~\cite{cai2023groot}. Each episode utilizes a unique environmental seed to ensure variability; however, all agents are tested with the same seed for consistency across episodes.
It is important to note that our experimental setup differs from that of STEVE-1~\cite{lifshitz2023steve} in that \textbf{\highlight{we initialize the agents in the biome most conducive to task completion}} for each task. 
Specifically, as mentioned in Supp.~\ref{supsub:Rules}, we utilize the ``\texttt{chat}'' action to initialize the agent. For the ``\texttt{Obtain diamonds}''~\raisebox{-0.3ex}{\includegraphics[width=0.3cm]{icon/diamond.png}} task, we equip the agent with night vision and a diamond pickaxe~\raisebox{-0.3ex}{\includegraphics[width=0.3cm]{icon/diamond_pickaxe.png}}, which is consistent with the description provided in the main paper of GROOT~\cite{cai2023groot}.
\textbf{\highlight{Considering that STEVE-1~\cite{lifshitz2023steve} may not be explicitly trained on the ``\texttt{mine horizontally}''~\raisebox{-0.3ex}{\includegraphics[width=0.3cm]{icon/iron_pickaxe.png}} instruction, we augment STEVE-1~\cite{lifshitz2023steve}'s prior original training data with the corresponding text-video pairs from the Goal Drift Dataset and retrain the prior}}. This ensures that the updated prior can map the textual instruction ``\texttt{mine horizontally}''~\raisebox{-0.3ex}{\includegraphics[width=0.3cm]{icon/iron_pickaxe.png}} to the associated visual instructions.
The detailed settings for the Command-Switching Evaluation for Long-Horizon Tasks experiment can be found in \cref{tab:Command-Switching Evaluation for Long-Horizon Tasks Datails}.

\begin{table*}[htbp]
\caption{\textbf{The detailed settings for the Command-Switching Evaluation.}} 
\centering
\small
\begin{tabular}{c|c|c|c|c|c}
\bottomrule[1pt]
Id & Text Instruction & Biome & Switch Condition   & Time Limit & Imagination Interval \\
\hline 
\multirow{2}{*}{1} & \texttt{chop a tree}~\raisebox{-0.4ex}{\includegraphics[width=0.3cm]{icon/tree.png}} & \multirow{2}{*}{Forest} & \multirow{2}{*}{Reach 1500 Frames} & \multirow{2}{*}{3000 Frames} & \multirow{2}{*}{25 Frames}\\
 & \texttt{craft wooden planks}~\raisebox{-0.3ex}{\includegraphics[width=0.3cm]{icon/plank.png}} &  &  &  &\\
\hline 
\hline 
\multirow{2}{*}{2} & \texttt{collect dirt}~\raisebox{-0.3ex}{\includegraphics[width=0.3cm]{icon/dirt.png}} & \multirow{2}{*}{Plains} & \multirow{2}{*}{Reach 2000 Frames}& \multirow{2}{*}{3000 Frames} & \multirow{2}{*}{25 Frames} \\
 & \texttt{build a tower}~\raisebox{-0.3ex}{\includegraphics[width=0.3cm]{icon/tower.png}} &  &  &  &\\
\hline 
\hline 
\multirow{2}{*}{3} & \texttt{dig down}~\raisebox{-0.3ex}{\includegraphics[width=0.3cm]{icon/iron_pickaxe.png}} & \multirow{2}{*}{Plains} & \multirow{2}{*}{Reach 13th floors}  & \multirow{2}{*}{12000 Frames}& \multirow{2}{*}{25 Frames}\\
 & \texttt{mine horizontally}~\raisebox{-0.3ex}{\includegraphics[width=0.3cm]{icon/iron_pickaxe.png}} &  &  &  &\\

% \arrayrulecolor{black}
\bottomrule[1pt]
\end{tabular}
\label{tab:Command-Switching Evaluation for Long-Horizon Tasks Datails}
\vspace{-4mm}
\end{table*}

\section{More Ablation Studies}
\label{sup:Ablation}

In this section, we introduce additional ablation studies to explore various contributors to performance. 
This includes the use of Classifier-Free Guidance~\cite{ho2022classifier} during inference, the selection of Drift Lengths from the Goal Drift Dataset, and the generation strategies for Visual Prompts. 
We employ the same experimental settings as our Programmatic Evaluation, compare the performance of different ablations, and plot the results, showing both the mean and 95\% confidence intervals of the programmatic metrics.

\subsection{Classifier-Free Guidance During Inference}

Given that VPT~\cite{baker2022video} is a foundation model obtained through behavior cloning from extensive video demonstrations without instruction guidance during training, this may lead to a smoother behavior distribution learned by VPT~\cite{baker2022video}. Consequently, even after fine-tuning VPT~\cite{baker2022video} for instruction-following abilities, when provided with direct instruction as a condition, it tends to act based on its previously learned knowledge from behavior cloning. It fails to steadily follow the instructions given, similar to the observation in Appendix I of Baker et al.~\cite{baker2022video}.
We believe that this bias arises inherently from the training process of VPT~\cite{baker2022video}. Inspired by STEVE-1~\cite{lifshitz2023steve}, we employ classifier-free guidance~\cite{ho2022classifier} to mitigate this bias as much as possible in the action logits space before sampling the action.
Specifically, for each inference, we perform two computations of logits through PolicyNet: one with visual prompt guidance and the other without. At each timestep, \textbf{\highlight{we subtract a certain proportion of the action logits from the unconditioned PolicyNet from those predicted by the visual prompt-conditioned PolicyNet}}. The equation for computing logits is directly borrowed from STEVE-1~\cite{lifshitz2023steve}.
{
\small
\begin{equation}
f_{t} \leftarrow \mathcal{V}(\mathcal{O}_{t}),~~~ o_{t} \leftarrow f_{t} + p_{t}, ~~~\operatorname{logits} \leftarrow (1+\lambda) \underbrace{\mathcal{T}_{\theta}\left(o_{t-T}, \ldots, o_{t}\right)}_{\text {conditional logits }}-\lambda \underbrace{\mathcal{T}_{\theta}\left(f_{t-T}, \ldots, f_{t}\right)}_{\text {unconditional logits }}
\tag{1}
\end{equation}
}

\noindent where $\mathcal{V}$ is the VisualEncoder and $\mathcal{T}$ is the TransformerXL~\cite{dai2019transformer}, $\mathcal{O}_t$ is the current observation, $p_t$ is the visual prompt, $\lambda$ is the trade-off parameter between the visual prompt conditioned logits and unconditioned logits.
By setting a suitable value for $\lambda$, we can encourage PolicyNet to follow the instructions in action generation more steadily.
\begin{figure*}[t]
\centering
\includegraphics[width=1\linewidth]{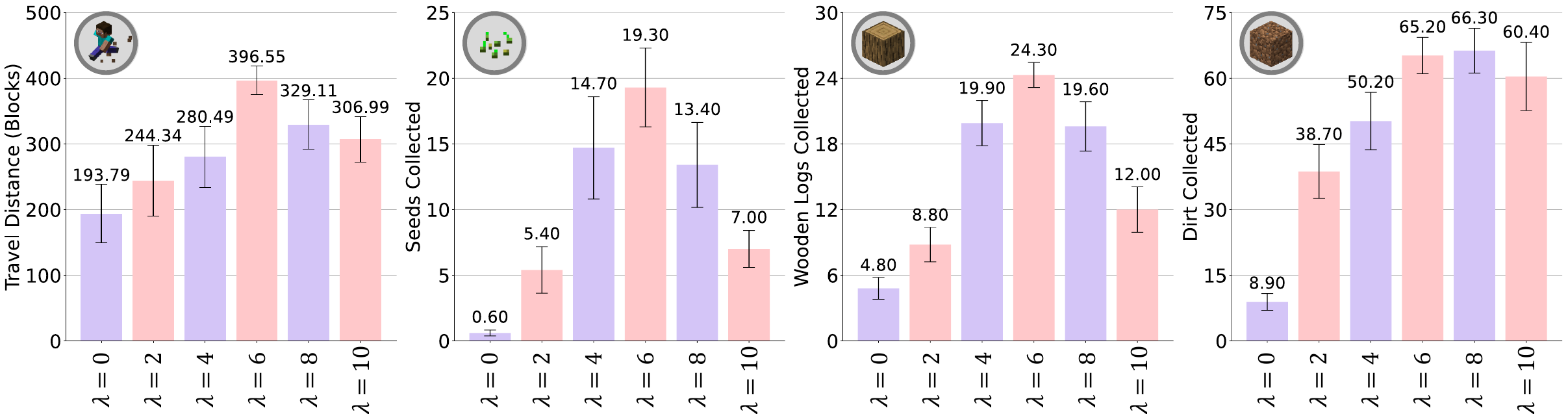}
   \caption{\textbf{The impact of different values of condition scale \( \lambda \) on the performance of the agent by using classifier-free guidance.} Selecting the optimal parameter \(\lambda\) to balance between visual prompt-conditioned and unconditioned settings can significantly enhance agent performance, consistently improving its ability to follow instructions. By using the best $\lambda$ ($\lambda$ = 6), {\mname} when significantly outperforms {\mname} when $\lambda$ = 0~(no guidance), collecting 32$\times$ more seeds~\raisebox{-0.3ex}{\includegraphics[width=0.3cm]{icon/seed.png}}, 5$\times$ more wood~\raisebox{-0.3ex}{\includegraphics[width=0.3cm]{icon/wood.png}}, 7.3$\times$ more dirt~\raisebox{-0.3ex}{\includegraphics[width=0.3cm]{icon/dirt.png}}, travelling 2$\times$ further~\raisebox{-0.3ex}{\includegraphics[width=0.3cm]{icon/explore.png}}.}
\vspace{-5.5mm}
\label{fig:Classifier-Free Guidance During Inference}
\end{figure*}
\cref{fig:Classifier-Free Guidance During Inference} illustrates how choosing different values affects the agent's performance in Programmatic Evaluation.
When the value of $\lambda$ is less than 6, performance improves with an increase in $\lambda$, indicating that classifier-free guidance~\cite{ho2022classifier} can significantly reduce the bias introduced by prior behavior. The agent performs optimally when $\lambda$ is 6 to 8; beyond this range, the performance begins to decline. This decrease is due to excessive guidance disrupting the agent's original understanding and knowledge of the environment, impeding its ability to act normally. 
Ultimately, we opt for a value of $\lambda$ equal to 6.
After utilising classifier-free guidance~\cite{ho2022classifier} during inference, {\mname} when $\lambda$ = 6 significantly outperforms {\mname} when $\lambda$ = 0~(no guidance), collecting 32$\times$ more seeds~\raisebox{-0.3ex}{\includegraphics[width=0.3cm]{icon/seed.png}}, 5$\times$ more wood~\raisebox{-0.3ex}{\includegraphics[width=0.3cm]{icon/wood.png}}, 7.3$\times$ more dirt~\raisebox{-0.3ex}{\includegraphics[width=0.3cm]{icon/dirt.png}}, travelling 2$\times$ further~\raisebox{-0.3ex}{\includegraphics[width=0.3cm]{icon/explore.png}}. 
Therefore, selecting an appropriate value for parameter $\lambda$ to balance the trade-off between visual prompt-conditioned and unconditioned logits can significantly enhance the agent's performance and steadily improve its ability to follow instructions in action generation.
Although this technique trick is effective during inference, it still needs to find the best hyperparameter in practice. In the future, eliminating biased behaviors directly from the fine-tuning process training would be meaningful.

\subsection{Selection of Drift Lengths}
During Goal Drift Dataset collection, we utilize fixed values for \( T_b \) and \( T_f \) to address the challenges of ``Goal Illusion'' and ``Imagination Stagnation'' by performing backward and forward drifts around the event occurrence moment \( t^* \). The specific algorithmic procedure is detailed in \cref{subsub:Goal Drift Dataset Collection} of the main paper.

\begin{figure*}[t]
\centering
\includegraphics[width=1\linewidth]{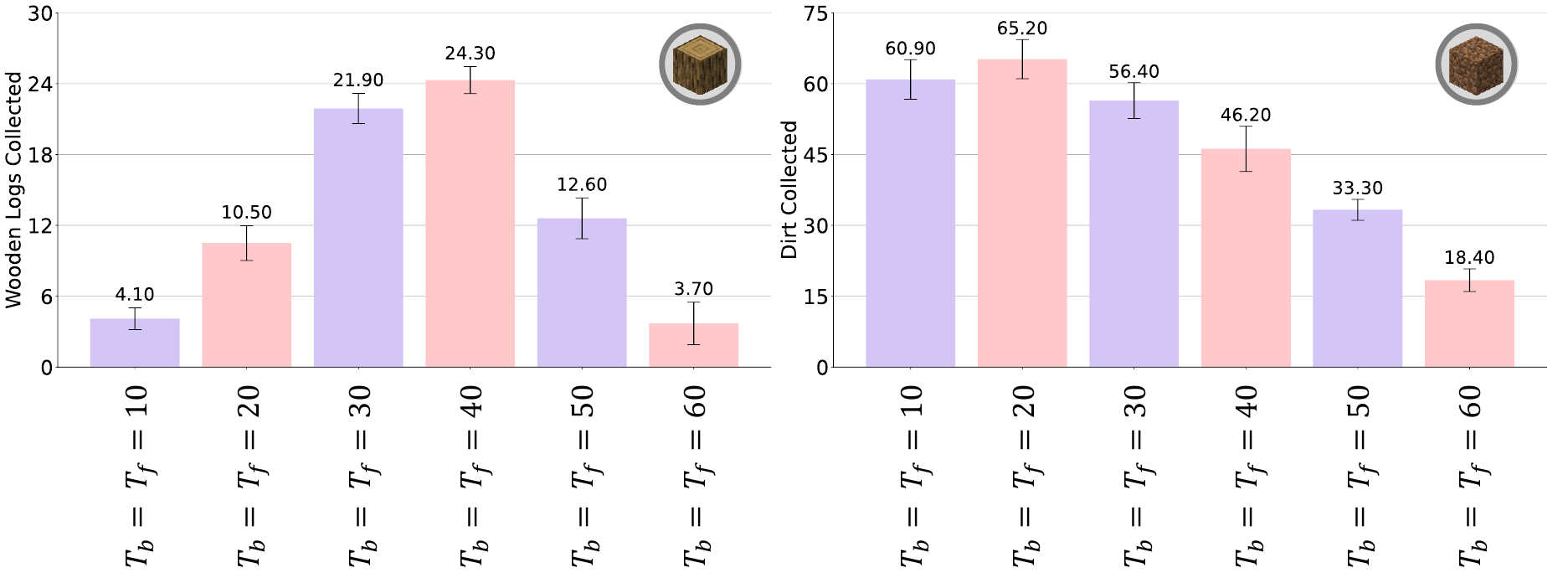}
   \caption{\textbf{The influence of different goal drift lengths on agent performance.} For each event, there is an optimal goal drift length that is correlated with the duration of an instruction task completion one time. Employing the appropriate goal drift length can address the dual challenges of ``Goal Illusion'' and ``Imagination Stagnation'', thereby enhancing the agent's ability to steadily follow instructions in action generation.}
\vspace{-5.5mm}
\label{fig:Selection of Drift Lengths}
\end{figure*}

It is noteworthy that we observe an inconsistency in the optimal Drift Length for each event. As illustrated in \cref{fig:Selection of Drift Lengths}, the optimal drift length for ``\texttt{wood}''~\raisebox{-0.3ex}{\includegraphics[width=0.3cm]{icon/wood.png}} is approximately 40, while for `\texttt{dirt}''~\raisebox{-0.3ex}{\includegraphics[width=0.3cm]{icon/dirt.png}}, it is around 20. We find that the best drift length correlates with the amount of time required to complete the instruction task once.
Also as shown in \cref{fig:Selection of Drift Lengths}, selecting appropriate values for \( T_b \) and \( T_f \) can effectively address the ``Goal Illusion'' and ``Imagination Stagnation'' challenges mentioned in \cref{subsub:Goal Drift Dataset Collection} of the main paper, enhancing the agent's ability to steadily follow instructions in action generation. Although this method is effective, it does require the cumbersome task of selecting the right length for each event. We believe that mitigating or eliminating the interference caused by varying drift lengths during training or fine-tuning in the future will be meaningful.

\subsection{Generation Strategies for Visual Prompts}
When the agent acts in the simulator, the Imaginator first creates a goal imagination of the next stage to complete the given instruction based on the current observation and instruction. Then, the Prompt Generator creates a visual prompt from this goal imagination, integrating the current observation and instruction. 
This part investigates strategies for generating visual prompts. We consider two variants:
\begin{enumerate}
    \item Unlike current methods, we can synthesize the imagination into a 16-frame video by simply stacking it 16 times, and we encode this video with the MineCLIP~\cite{fan2022minedojo} video encoder to project it into the MineCLIP~\cite{fan2022minedojo} space and align it with the PolicyNet using a linear layer. More specifically, we bypass the reconstruction step mentioned in Supp.~\ref{supsec:Prompt Generator} and directly transform the only goal imagination into the required visual prompt for the MineCLIP~\cite{fan2022minedojo} visual space.
    \item In contrast to current methods, we eliminate the Imaginator and retrain a Prompt Generator to directly reconstruct visual prompts from current observations and instructions. Specifically, we retrain a CVAE~\cite{sohn2015learning, kingma2013auto} without using the goal of Imagination as a guiding condition for visual prompt generation.
\end{enumerate}
\begin{figure*}[t]
\centering
\includegraphics[width=1\linewidth]{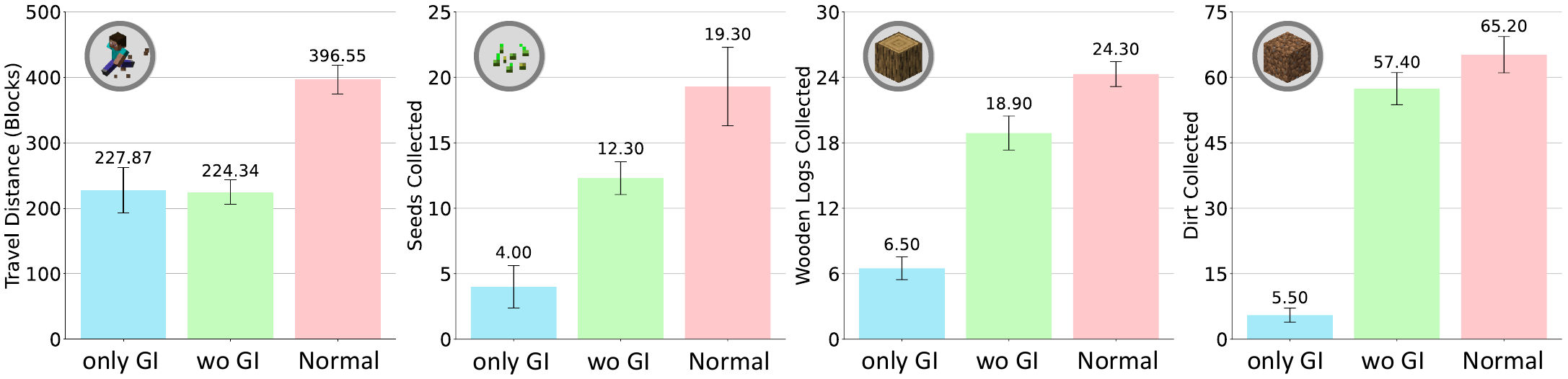}
   \caption{\textbf{The impact of different visual prompt generation strategies on the performance.} ``only GI'' refers to bypassing the CVAE reconstruction phase in Prompt Generator and directly stacking the goal imagination into a static 16-frame video as the visual prompt.  ``wo GI'' indicates that the CVAE reconstructs the visual prompt without using goal imagination as a condition, thus skipping the imagination phase of the Imaginator.}
\vspace{-5.5mm}
\label{fig:Generation Strategies for Visual Prompts}
\end{figure*}

From \cref{fig:Generation Strategies for Visual Prompts}, it is evident that our approach enables the agent to follow instructions more steadily, as our visual prompt provides a more precise demonstration of the desired behaviour customized to the current environment.
One drawback of using goal imagination stacked into a 16-frame video as a visual prompt is that the depicted behavior resembles a static state. This can confuse PolicyNet, making it unclear whether to remain stationary or to achieve the state represented in the video.
A limitation of reconstructing the current observation and instruction into a visual prompt is that the CVAE~\cite{sohn2015learning, kingma2013auto}'s ability to model future spatiotemporal aspects is subpar. Without relying on goal imagination, it struggles to accurately reconstruct the demonstration of the desired behaviour. This occasionally results in misleading the agent, preventing it from steadily following instructions during action generation.

\section{More Visual Results}
\label{sup:Visual}
Imagination visual results are all hosted on our \href{https://sites.google.com/view/minedreamer/main}{project webpage}.

\subsection{Imagination Visual Results without Goal Drift}
\begin{figure*}[htbp]
\centering
\includegraphics[width=1\linewidth]{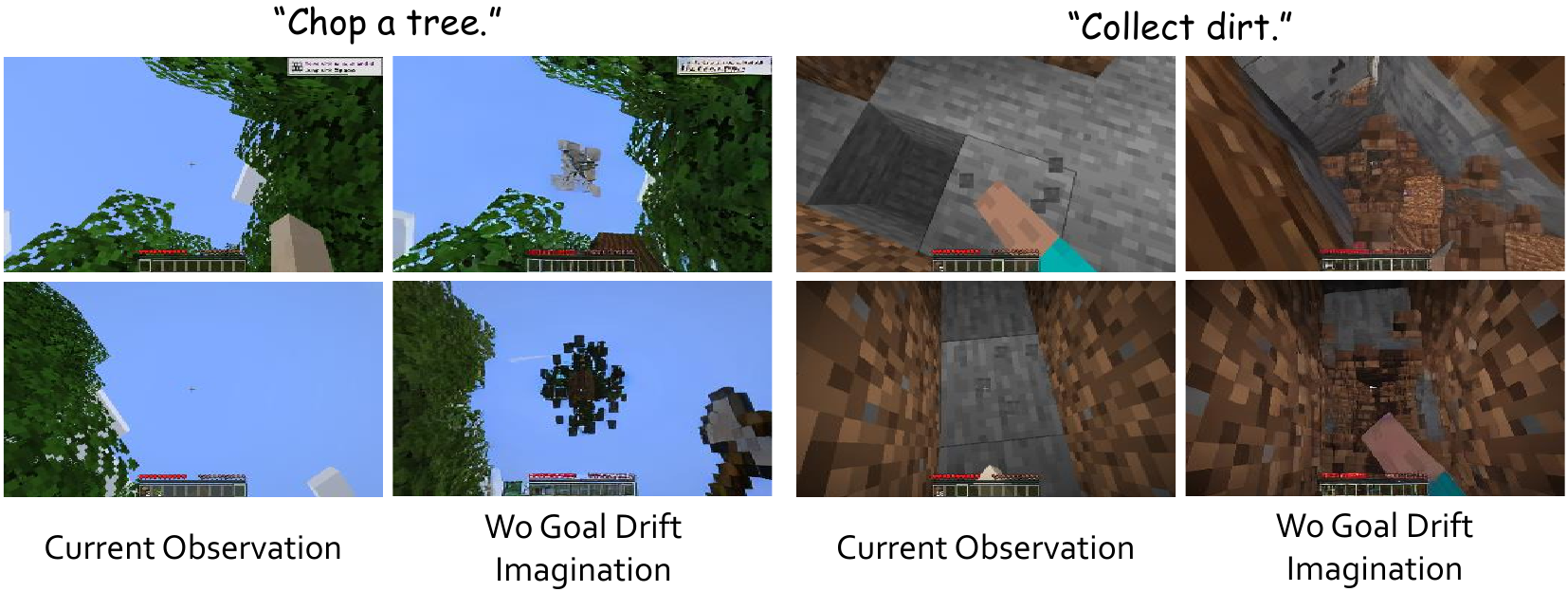}
% \vspace{-5.5mm}
   \caption{\textbf{Imagination Visual Results without Goal Drift.} Due to the absence of goal drift, the imaginations generated by the Imaginator are all related to the moment of event-related instruction completion, leading to the phenomenon known as ``Goal Illusion'', where the Imaginator edits the current observation to represent the executed instruction. In the figure depicted, the agent inserts broken wooden blocks~\raisebox{-0.3ex}{\includegraphics[width=0.3cm]{icon/wood.png}} into the sky and, facing a stone~\raisebox{-0.3ex}{\includegraphics[width=0.3cm]{icon/stone.png}}, imagines itself breaking dirt~\raisebox{-0.3ex}{\includegraphics[width=0.3cm]{icon/dirt.png}}.}
% \vspace{-5.5mm}
\label{fig:wo backward drift}
\end{figure*}

\begin{figure*}[htbp]
\centering
\includegraphics[width=1\linewidth]{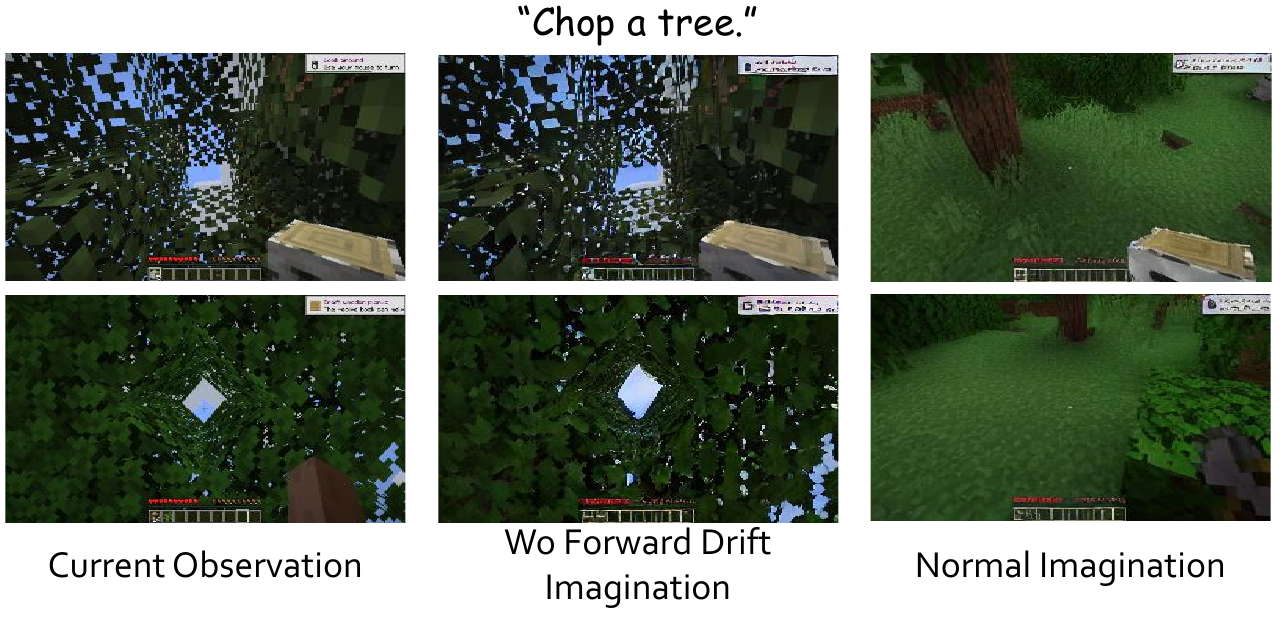}
% \vspace{-5.5mm}
   \caption{\textbf{Imagination Visual Results without Forward Drift.} Due to the lack of forward drift, the imaginations produced by the Imaginator are all from moments prior to the completion of event-related instructions, resulting in a phenomenon called ``Imagination Stagnation''. This means the Imaginator fails to anticipate the outcomes of repeated tasks. For example, in the figure provided, after the agent cuts the uppermost wood~\raisebox{-0.3ex}{\includegraphics[width=0.3cm]{icon/wood.png}} by looking up, it will not look down for more trees~\raisebox{-0.4ex}{\includegraphics[width=0.3cm]{icon/tree.png}} to continue the task.}
% \vspace{-5.5mm}
\label{fig:wo forward drift}
\end{figure*}

To evaluate the efficacy of the Goal Drift data collection method, we carry out experiments comparing various data collection approaches. 
\cref{fig:wo backward drift} illustrates the imagination generated by the Imaginator trained on data collected without any goal drift. Due to the absence of backward drift, all imaginations generated by the Imaginator correspond to the moment when the event-related instructions are completed. Consequently, this leads to the phenomenon of ``Goal Illusion'', where the Imaginator edits the current observation to depict the completed instruction. For the instruction ``\texttt{Chop a tree}''~\raisebox{-0.4ex}{\includegraphics[width=0.3cm]{icon/tree.png}}, when the agent faces the sky, the Imaginator may unrealistically insert a broken wooden log~\raisebox{-0.3ex}{\includegraphics[width=0.3cm]{icon/wood.png}} into the sky. For the instruction ``\texttt{Collect dirt}''~\raisebox{-0.3ex}{\includegraphics[width=0.3cm]{icon/dirt.png}}, even though the agent is pointing at a stone~\raisebox{-0.3ex}{\includegraphics[width=0.3cm]{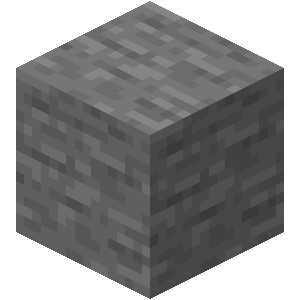}}, the Imaginator still imagines dirt and shatters it, resulting in the agent eventually attempting to break the stone~\raisebox{-0.3ex}{\includegraphics[width=0.3cm]{icon/stone.png}}.
\cref{fig:wo forward drift} shows the imaginations generated by the Imaginator trained on data collected without forward drift. Because there is no forward drift, all imaginations generated by the Imaginator represent moments before the completion of event-related instructions. This results in the phenomenon of ``Imagination Stagnation'',  where the Imaginator fails to conceive repeated task completion.
For the instruction ``\texttt{Chop a tree}''~\raisebox{-0.4ex}{\includegraphics[width=0.3cm]{icon/tree.png}}, after cutting the uppermost wood~\raisebox{-0.3ex}{\includegraphics[width=0.3cm]{icon/wood.png}} by looking up, the agent will not look down for more trees~\raisebox{-0.4ex}{\includegraphics[width=0.3cm]{icon/tree.png}}, which impedes continuous task performance. In contrast, an Imaginator trained with data collected including forward drift is able to understand that the agent should now look down to find other trees~\raisebox{-0.4ex}{\includegraphics[width=0.3cm]{icon/tree.png}} to continue the task.

\subsection{Imagination Visual Results on Evaluation Set}
\begin{figure*}[htbp]
\centering
\includegraphics[width=1\linewidth]{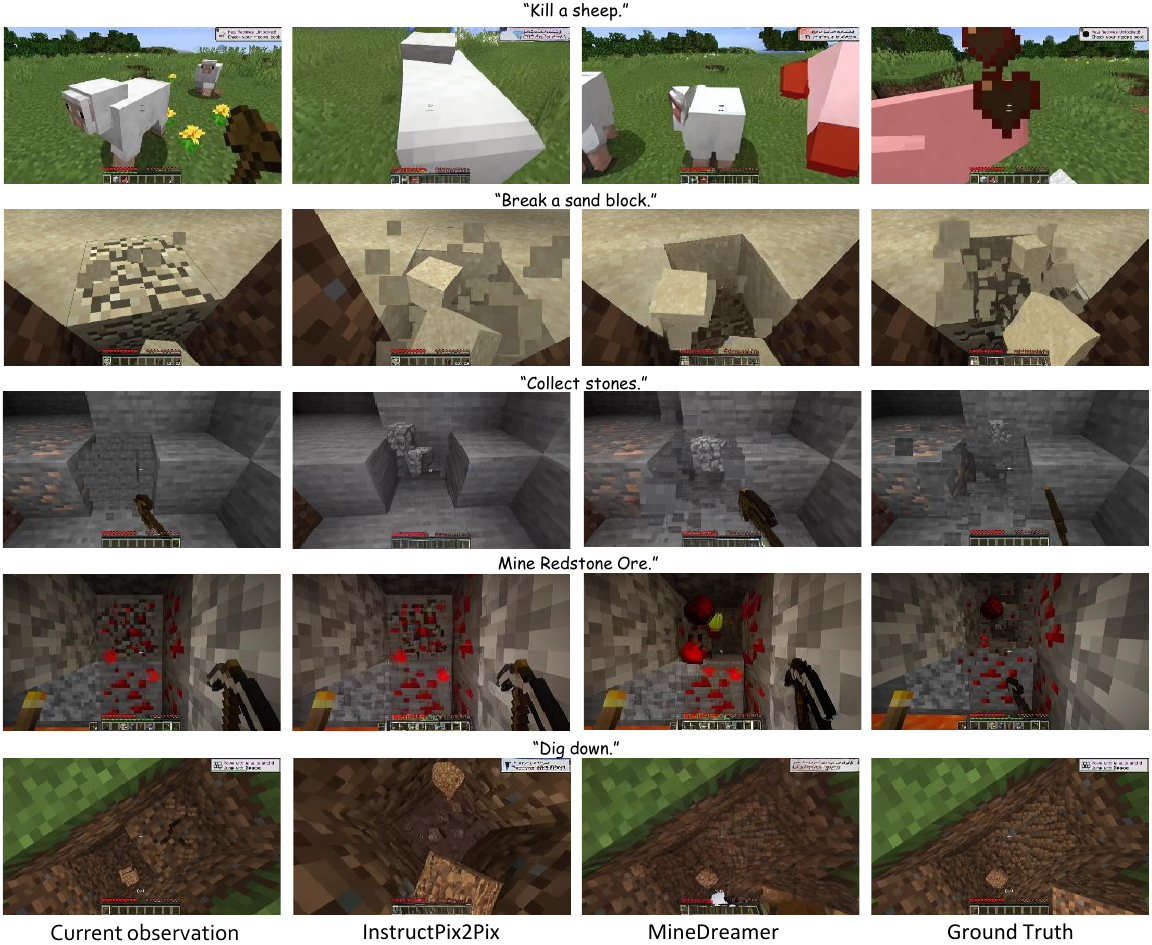}
% \vspace{-5.5mm}
   \caption{Imagination visual results on Goal Drift Evaluation Set.}
% \vspace{-5.5mm}
\label{fig:evaluation in sup}
\end{figure*}

We compare Imaginator with the existing state-of-the-art instruction-based image editing model, namely InstructPix2Pix~\cite{brooks2023instructpix2pix}. Given this model has been trained on specific datasets, its performance would inevitably be suboptimal if directly applied to the Minecraft domain. To facilitate a fair comparison, we fine-tune InstructPix2Pix~\cite{brooks2023instructpix2pix} using the same training set employed by the Imaginator and evaluate the performance of the fine-tuned models in addressing tasks in Minecraft. Fig~\ref{fig:evaluation in sup} shows qualitative results in the evaluation set, our methodology exhibits enhanced abilities in Goal Imagination Generation within intricate scenarios.

\subsection{Imagination Visual Results During Agent Solving Tasks}
\begin{figure*}[htbp]
\centering
\includegraphics[width=1\linewidth]{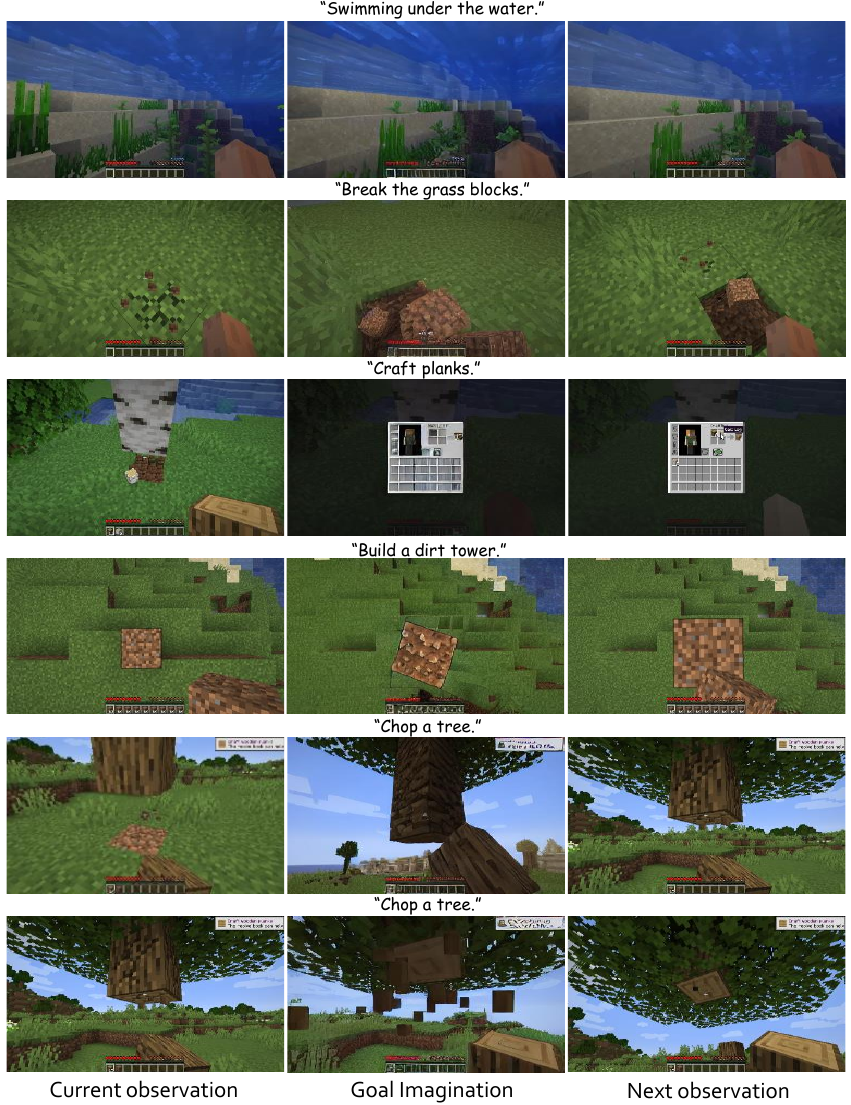}
% \vspace{-5.5mm}
   \caption{Imagination visual results during agent solving tasks.}
% \vspace{-5.5mm}
\label{fig:inference in sup}
\end{figure*}

\begin{figure*}[htbp]
\centering
\includegraphics[width=1\linewidth]{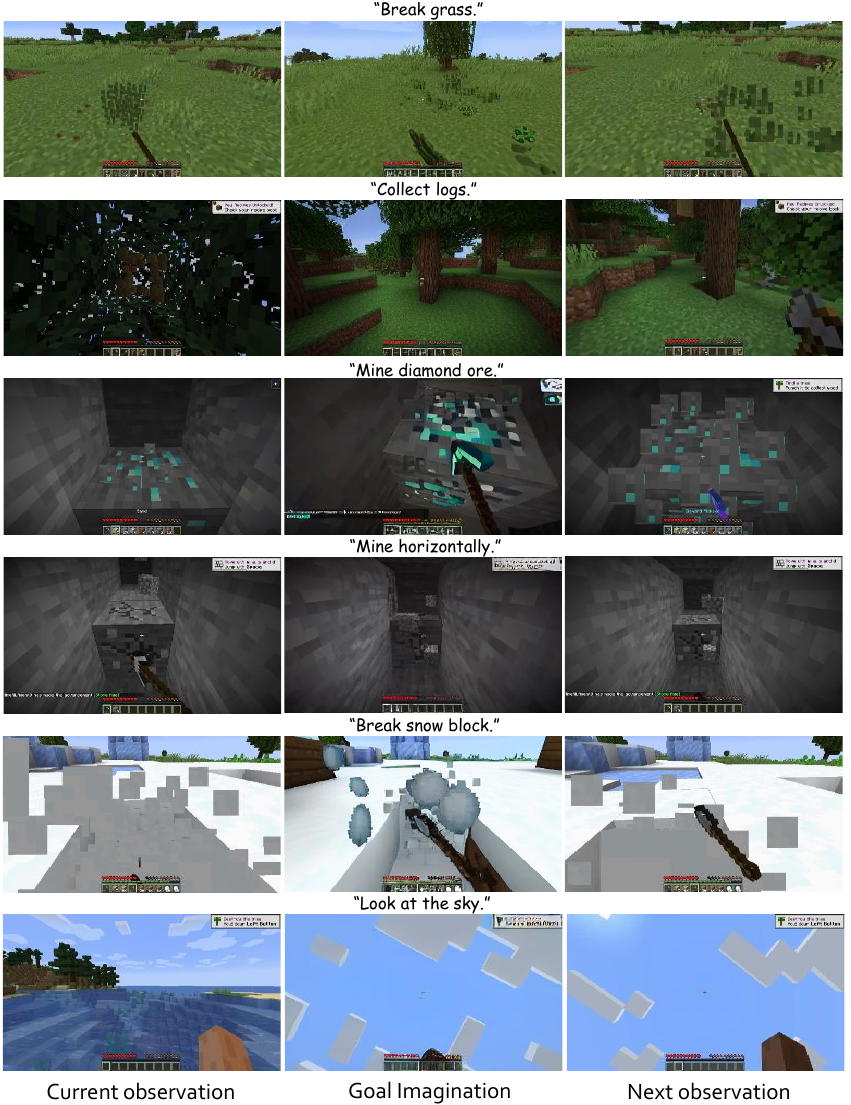}
% \vspace{-5.5mm}
   \caption{Imagination visual results during agent solving tasks.}
% \vspace{-5.5mm}
\label{fig:inference_1 in sup}
\end{figure*}

We visualize the agent's imagination during task execution alongside the next observation in Fig.~\ref{fig:inference in sup} and Fig.~\ref{fig:inference_1 in sup} to evaluate the Imaginator's generalization capability in open scenarios. It is observed that the Imaginator is capable of generating high-quality visualizations that closely align with the current scene in an open environment, thereby guiding the subsequent PolicyNet to autoregressively predict the next action steadily.

\subsection{User Studies}
\begin{figure*}[htbp]
\centering
\includegraphics[width=1\linewidth]{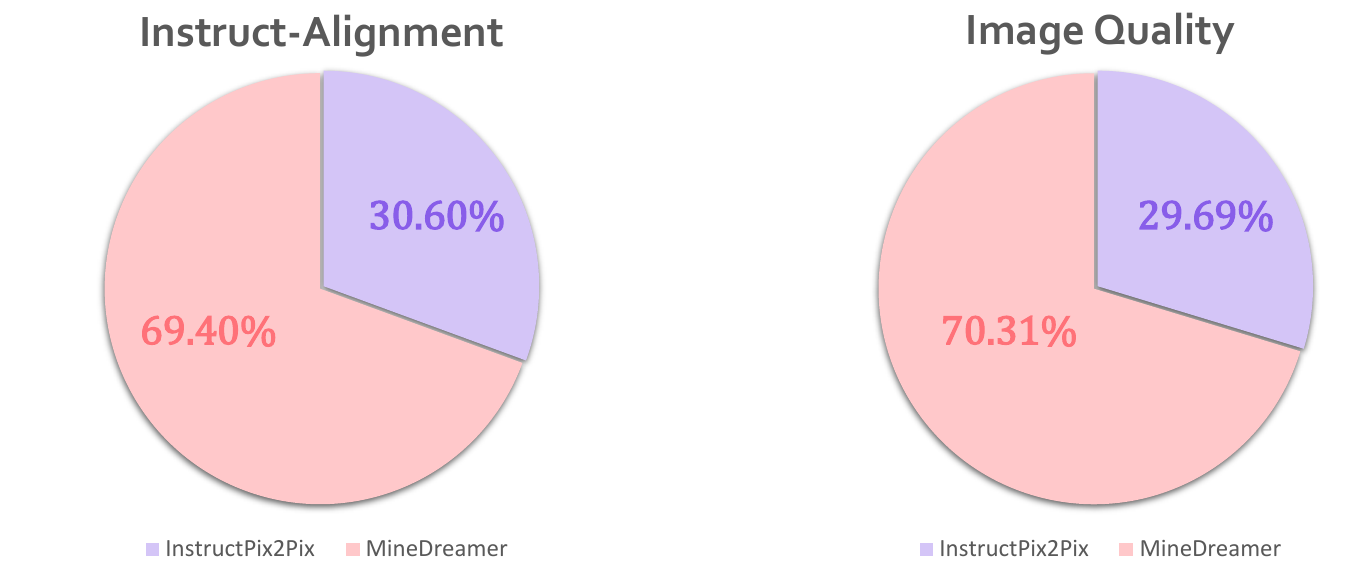}
% \vspace{-5.5mm}
   \caption{\textbf{The results of user studies}, comparing the results generated by InstructPix2Pix and \mname. Based on the results from both the Instruction Alignment and Image Quality perspectives, \mname  demonstrates superior effectiveness.}
% \vspace{-5.5mm}
\label{fig:user study in sup}
\end{figure*}

To further evaluate \mname's efficacy, we conduct a user study. Specifically, we randomly select 15 images from the evaluation set, representing a wide range of tasks and scenarios within Minecraft. For each image, we generate results using both InstructPix2Pix~\cite{brooks2023instructpix2pix} and \mname, then randomly shuffle the order of these results. As noted in \cref{sub: Qualitative Results of Imaginator} of the main paper, InstructPix2Pix~\cite{brooks2023instructpix2pix} is fine-tuned on the same dataset as \mname. This process yield 15 sets of images in a shuffled sequence. Participants are asked to independently identify the two superior images for each set: the first being the one that best matches the given instructions (named \textbf{\highlight{Instruct-Alignment}}), and the second being the image that most closely mirrors real-world appearances, including perspective and physical laws (named \textbf{\highlight{Image Quality}}). A total of 25 individuals participate in the study. The findings, illustrated in Fig.~\ref{fig:user study in sup}, reveal that over $69.40\%$ of participants find \mname's outputs to be more aligned with the instructions, and more than $70.31\%$ favor the results produced by \mname for their realism. These outcomes further underscore \mname's instruction following ability and generalization ability. 

\section{Demo Videos}
\label{sup:Videos}

Demo videos are all hosted on our \href{https://sites.google.com/view/minedreamer/main}{project webpage}.

\subsection{Programmatic Evaluation}
We demonstrate videos of the four tasks from the Programmatic Evaluation on the aforementioned anonymous project webpage. Of course, you can also view the demo videos for the respective tasks by directly accessing the video URLs.

\begin{itemize}
    \item ``\texttt{Go explore}''~\raisebox{-0.3ex}{\includegraphics[width=0.3cm]{icon/explore.png}}: \url{https://youtu.be/UdG0ckoGRCY}
    \item ``\texttt{Collect seeds}~\raisebox{-0.3ex}{\includegraphics[width=0.3cm]{icon/seed.png}}'': \url{https://youtu.be/TFchu_YBiuI}
    \item ``\texttt{Chop a tree}''~\raisebox{-0.4ex}{\includegraphics[width=0.3cm]{icon/tree.png}}: \url{https://youtu.be/Sx_NKjq5DTA}
    \item ``\texttt{Collect dirt}''~\raisebox{-0.3ex}{\includegraphics[width=0.3cm]{icon/dirt.png}}: \url{https://youtu.be/7TOR0SOFaB8}
\end{itemize}

\subsection{Command-Switching Evaluation}
We demonstrate videos of the three tasks from the Command-Switching Evaluation on the anonymous project webpage mentioned above. Of course, you can also view the demo videos for the respective tasks by accessing the video URLs.

\begin{itemize}
    \item ``\texttt{Chop a tree~\raisebox{-0.4ex}{\includegraphics[width=0.3cm]{icon/tree.png}} to Craft planks~\raisebox{-0.3ex}{\includegraphics[width=0.3cm]{icon/plank.png}}}'': \url{https://youtu.be/YtY2M_Hi7OE}
    \item ``\texttt{Gather dirt~\raisebox{-0.3ex}{\includegraphics[width=0.3cm]{icon/dirt.png}} to Build a tower~\raisebox{-0.3ex}{\includegraphics[width=0.3cm]{icon/tower.png}}}'': \url{https://youtu.be/Zy2t2RpeNtQ}
    \item ``\texttt{Obtain Diamond}''~\raisebox{-0.3ex}{\includegraphics[width=0.3cm]{icon/diamond.png}}: \url{https://youtu.be/hThbWh0q5EE}

\end{itemize}

\end{document}